\PassOptionsToPackage{dvipsnames}{xcolor}
\documentclass[letterpaper,10 pt,conference]{ieeeconf}  
\IEEEoverridecommandlockouts             
\usepackage{graphicx} 
\usepackage{cite} 
\usepackage{multicol}
\usepackage{graphicx}
\usepackage{subcaption}
\usepackage{subfiles}
\usepackage{multirow}
\usepackage[many]{tcolorbox}
\usepackage{hyperref}
\usepackage{cleveref}

\graphicspath{{figs/}{../figs/}}

\let\labelindent\relax

\usepackage[inline]{enumitem}

\usepackage{xspace}
\def\flow{RMP{flow}\xspace}

\newenvironment{flushitemize}{%
	\begin{list}{$\bullet$}
		{\setlength{\leftmargin}{15pt}}%
		\setlength{\labelwidth}{20pt}
		\setlength{\itemindent}{0pt}
		\setlength{\labelsep}{0.5em}
		\setlength{\itemsep}{1pt}
		\setlength{\parskip}{0pt}
		\setlength{\parsep}{0pt}}
	{\end{list}}

\usepackage{amsmath}
\usepackage{amsfonts}
\usepackage{amssymb}
\usepackage{amsthm}
\usepackage{bm}
\usepackage{bbm}
\usepackage{mathtools}
\usepackage{enumitem}
\usepackage{thmtools,thm-restate}
\usepackage{algorithm}
\usepackage{algorithmic}
\usepackage{graphicx}
\usepackage{comment}

\theoremstyle{plain}

\newtheorem{theorem}{Theorem}[section]

\theoremstyle{definition}

\theoremstyle{remark}

\newenvironment{proofsketch}{\textit{Proof sketch:}}{\qed \par}

\def\CC{\mathcal{C}}

\def\LL{\mathcal{L}}
\def\MM{\mathcal{M}}

\def\TT{\mathcal{T}}

\def\Lb{\mathbf{L}}
\def\Mb{\mathbf{M}}

\def\lb{\mathbf{l}}

\def\pb{\mathbf{p}}
\def\ub{\mathbf{u}}
\def\vb{\mathbf{v}}\def\wb{\mathbf{w}}

\def\Rbb{\mathbb{R}}

\def\dtt{\mathtt{d}}\def\ett{\mathtt{e}}

\def\ltt{\mathtt{l}}

\def\rtt{\mathtt{r}}
\def\utt{\mathtt{u}}
\def\vtt{\mathtt{v}}

\def\R{\Rbb}

\def\t{\top}
\def\*{\star}

\DeclareMathOperator*{\argmin}{arg\,min}

\newcommand{\p}{\mathbf{p}}
\newcommand{\q}{\mathbf{q}}
\newcommand{\qd}{{\dot{\q}}}

\newcommand{\x}{\mathbf{x}}

\newcommand{\y}{\mathbf{y}}

\newcommand{\z}{\mathbf{z}}

\newcommand{\J}{\mathbf{J}}
\newcommand{\Jd}{{\dot{\J}}}

\newcommand{\M}{\mathbf{M}}

\newcommand{\sdot}[2]{\overset{\lower0.1em\hbox{$\scriptscriptstyle #2$}}{#1}}

\title{\LARGE \bf Towards Coordinated Robot Motions: End-to-End Learning of Motion Policies on Transform Trees}

\author{M. Asif Rana$^{*1,3}$, Anqi Li$^{*2,3}$, Dieter Fox$^{2,3}$, Sonia Chernova$^{1}$, Byron Boots$^{2,3}$, and Nathan Ratliff$^{3}$
\thanks{$^*$ Indicates equal contribution. $^{1}$ Georgia Institute of Technology. $^{2}$ University of Washington. $^{3}$ NVIDIA Research}%
}

\begin{document}

\maketitle
\thispagestyle{empty}
\pagestyle{empty}

\begin{abstract}
    Generating robot motion that fulfills multiple tasks simultaneously is challenging due to the geometric constraints imposed by the robot. In this paper, we propose to solve multi-task problems through learning structured policies from human demonstrations. Our structured policy is inspired by RMPflow, a framework for combining subtask policies on different spaces. 
    The policy structure provides the user an interface to 1) specifying the spaces that are directly relevant to the completion of the tasks, and 2) designing policies for certain tasks that do not need to be learned. We derive an end-to-end learning objective function that is suitable for the multi-task problem, emphasizing the deviation of motions on task spaces. Furthermore, the motion generated from the learned policy class is guaranteed to be stable. 
    We validate the effectiveness of our proposed learning framework through qualitative and quantitative evaluations on three robotic tasks on a 7-DOF Rethink Sawyer robot.
\end{abstract}

\section{Introduction}\label{sec:introduction}
Robotic systems often need to consider multiple tasks simultaneously to achieve their overall missions. For example, consider the task of placing an object on a shelf. The end-effector of the robot needs to reach a goal location, while the whole body of the robot is required to
avoid collisions with shelf. As can be seen from the previous example, each task can be more conveniently described in its corresponding space, e.g. it is easier to specify the collision avoidance behavior in a 1-d distance space. 
Generating a motion that fulfills all tasks simultaneously is challenging, as the execution of each task is not independent due to the geometric constraints of the robot. 

Recently, a framework, called \flow~\cite{cheng2018rmpflow}, has been proposed for solving the aforementioned multi-task problem. \flow generates robot motion by combining task policies defined on different (and potentially correlated) spaces. It provides each task with an acceleration policy and a state-dependent importance weight matrix, the two of which together are called a Riemannian motion policy (RMP)~\cite{ratliff2018riemannian}. \flow then combine these policies through a message passing algorithm that solves a weighted least-squares problem defined by the importance weight matrices~\cite{cheng2020efficient}. 
It has been shown in~\cite{cheng2018rmpflow} that, when the RMPs satisfy certain geometric conditions, the motion generated by~\flow is Lyapunov stable. Due to its stability properties and computational efficiency, \flow has be applied to a variety of robotic systems for generating complex motions in multi-task setting, e.g. ~\cite{kappler2018real,li2019stable,li2019multi,meng2019neural,wingo2020extending}.

Despite its rich expressivity, it is in general hard to design an RMP: it requires designing the state-dependent importance weight matrix, which for the stability properties to hold, also adds complications for designing the policies. One way to overcome the diffculty of designing RMPs is through learning RMPs from human demonstrations, in particular, kinesthetic teaching, as it provides the user an intuitive way communicate with the robot the desired behaviors.

\begin{figure}
    \centering
    \includegraphics[trim=140 10 40 10, clip, width=0.8\linewidth]{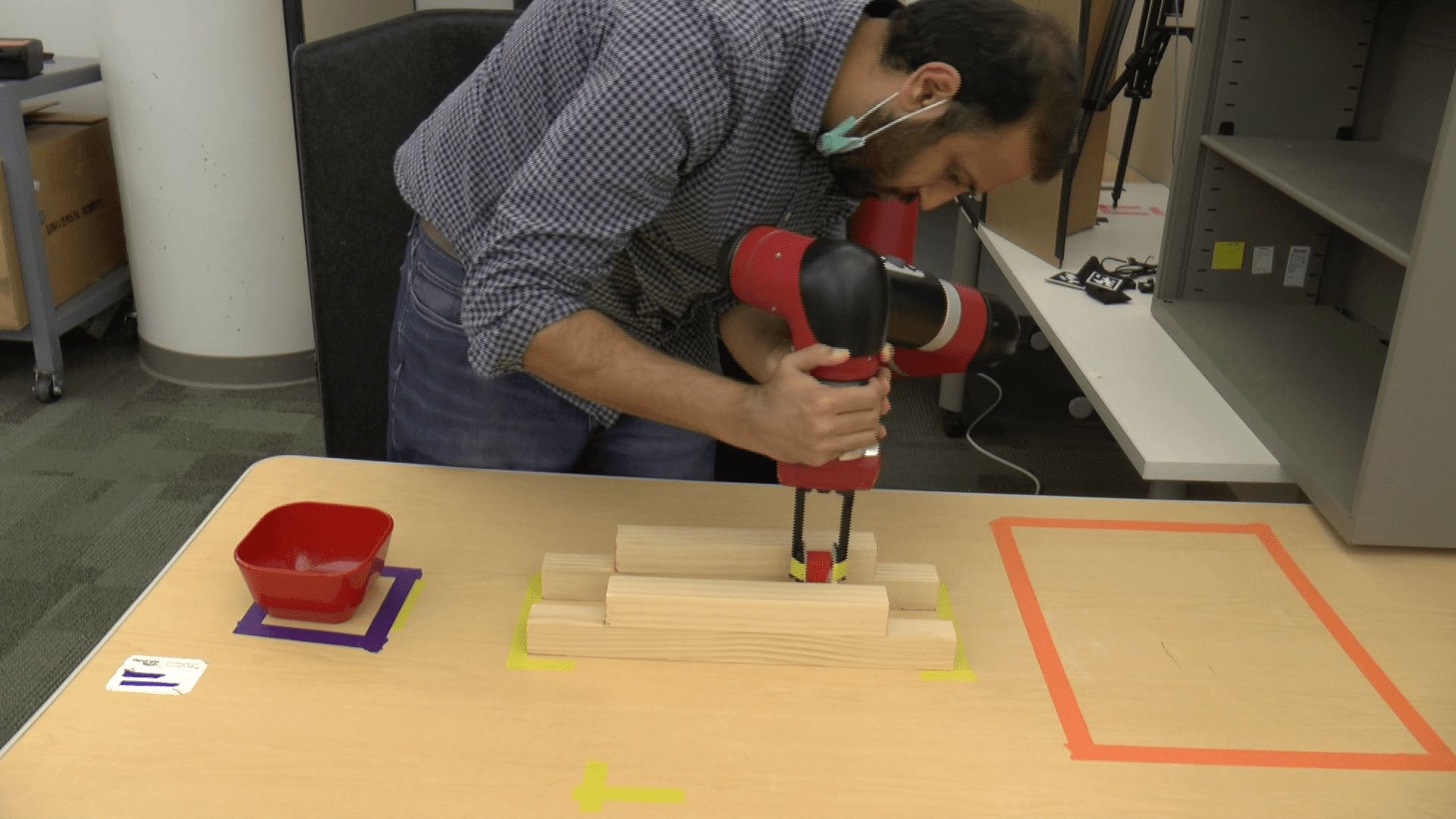}
    \caption{{A human providing demonstrations to the robot for a manipulation task through kinesthetic teaching.}}
    \label{fig:lfd}
\end{figure}

Using the \flow structure for learning from demonstrations (LfD) has the following benefits.
First, the learned motion is guaranteed to be stable as long as all task RMPs 
are properly designed or parameterized. 
Second, it provides the user a convenient interface to specifying which spaces are relevant to the tasks. For instance, if the end-effector position is what only matters to the tasks, the human may provide joint space trajectories that seem conflicting with one another (although they are consistent when viewed in the workspace). Directly regressing on the joint space trajectories will produce large error both in the joint space and in the workspace. 
Lastly, it also allows policies to be hand-designed for certain tasks, e.g. joint damping, redundancy reslution, collision avoidance, while the other policies are learned from data. This provides the user with freedom in deciding which task policies should be learned.

Existing work~\cite{rana2020learning} has explored learning RMPs from human demonstrations. 
In~\cite{rana2020learning}, each RMP is \textit{independently} learned to reproduce the demonstrated trajectories mapped to the corresponding task space. 
After learning, these independently learned RMPs, as well as hand-designed ones, are combined together by the \flow algorithm to produce the configuration space policy. 
The major limitation of this work is that the learned importance weight matrices are not learned to provide the proper trade-offs between policies, as each policy is learned independently. Due to this, additional manual scaling of the importance weight matrices is needed especially when combined with hand-designed policies. Second, the geometric constraints (e.g. induced by the kinematics of the robot) between tasks are not considered during learning. 
Lastly,~\cite{rana2020learning} uses a policy parameterization with limited capacity. In summary, despite its empirical success, the approach introduced in~\cite{rana2020learning} is not able to fully exploit the benefits provided by the \flow structure. 
We will further demonstrate this in the experiments.

In this paper, we propose a velocity-control\footnote{This is practical as modern robots often have a reasonably-tuned low-level tracking control, allowing position and velocity-based control interfaces.} motion generation algorithm similar to \flow. We show that the new velocity-control formulation enjoys all the benefits of \flow mentioned above, while providing a simpler structure for parameterizing stable policies. 
Furthermore, we adopt a principled way in learning structured policies from human-demonstrations in an end-to-end fashion. In contrast to~\cite{rana2020learning}, during learning, we jointly consider all RMPs being learned as well as the hand-specified ones. A new objective function is proposed to characterize the deviation between the human demonstration and the learned overall policy measured in all task spaces with learable RMPs. We differentiate through the least-squares optimization procedure induced by the proposed \flow-type so that the geometric constraints between task spaces are accounted during learning. 
Finally, we incorporate an experssive parameterization of RMPs through learning a latent space policy, which is inspired a recent work in learning diffeomorphism~\cite{rana2020euclideanizing}.


\section{Related Work}\label{sec:related_work}
\textbf{Motion Generation for Multi-Task Problems:} 
A general strategy for solving this multi-task motion generation problem is to generate (through either designing or learning) controllers or policies for each task independently, and then provide a high-level rule to combine them. 
Null-space or hierarchical operational control assigns priorities to the tasks, and only allows the lower-priority policies to act on the null space of high-priority tasks~\cite{peters2008unifying,escande2014hierarchical,dietrich2012continuous,lee2012intermediate}. However, these approaches could suffer from algorithmic singularities, due to multiple projections, that may arise when there are a large number of tasks. If this occurs, the system can easily become unstable~\cite{dietrich2018hierarchical}. Instead of assigning priorities, \flow provide each task with a state-dependent importance weight matrix, and the motion is generated through solving a weighted least-squares problem defined by the importance weight matrix~\cite{cheng2020efficient}. 
As is discussed in~Section \ref{sec:introduction}, the motion generated by~\flow is Lyapunov stable as long as all RMPs follow certain geometric structure. 

\textbf{Learning from Human Demonstrations: }
Several approaches exist which seek to learn policies from human demonstrations. These methods are typically grouped into two categories: 1) time-dependent policy learning~\cite{paraschos2017probabilistic, pastor2009learning}, and 2) time-invariant policy learning~\cite{khansari2011learning,neumann2015learning,chaandar2019learning, rana2020euclideanizing}. As elaborated in~\cite{khansari2011learning}, time-dependent methods, including the well-known dynamic movement primitives~\cite{pastor2009learning, ijspeert2013dynamical}, are susceptible to fail when either the environment or the time-horizon of motions is dynamic. On the other hand, time-invariant policies, in the absence of stability guarantees, are likely to suffer from the compounding error problem~\cite{shimodaira2000improving}. Most previous approaches for learning stable time-invariant policies~\cite{khansari2011learning,neumann2015learning,chaandar2019learning,rana2020euclideanizing}, however, are limited to learning motions associated with a single task assigned to a given robot body part (e.g. center of the end-effector). 

\textbf{Learning Riemannian Motion Policies: }
Recent works have explored learning RMPs from data~\cite{meng2019neural,rana2020learning,mukadam2020riemannian,aljalbout2020learning}.  
Meng et al. \cite{meng2019neural} learn to map perception input to RMPs through imitating hand-designed RMPs in an autonomous navigation setting. However, the learned RMPs does not satisfy the geometric condition for generating stable motions. To fine-tune the motion generated by fixed hand-designed policies, Mukadam et al. \cite{mukadam2020riemannian} add learnable scalar weights to the \flow algorithm. The expressively of this policy class, however, is limited by the fixed, hand-designed policies. Aljalbout et al.~\cite{aljalbout2020learning} propose to learn collision avoidance RMPs through reinforcement learning, although the learned policy is not guaranteed to be stable. 

The work most relevant to this paper is~\cite{rana2020learning}, where it also learns RMPs from human demonstrations. To provide stability guarantees to the learned policy, it incorporates a neural network architecture to ensure the positive-definiteness of the importance weight matrix. However, as is mentioned in Section \ref{sec:introduction}, this work has the limitation due to the fact that the policies are learned {independently}, and also that the policy parameterization has limited capacity. 

\section{Motion Generation with Transform Trees}\label{sec:background}
In this section, we propose a new motion generation for velocity-based motion control inspired by~\flow~\cite{cheng2018rmpflow}. In Section III.B, we introduce the optimization problem for the velocity-based control problem. 
We then introduce our proposed algorithm in Section III.C and analyze the stability property of the algorithm in Section III.D. 

\subsection{Motion Generation for Multi-task Problems}
The goal of motion generation is to provide a configuration space trajectory given the desired behaviors on the task space. We consider multi-task problems, where the robot can be tasked with multiple specifications, which we call \emph{subtasks}, simultaneously. For example, consider the task of placing an object on a shelf. The end-effector of the robot needs to reach a goal location, while the whole body of the robot is required to
avoid collisions with shelf. A subtask can sometimes be more easily specified on its individual space, rather than the joint configuration space. For example, collision avoidance can be described as a behavior on the 1-dimensional distance field. This yields a motion generation problem with subtasks defined on different \emph{subtask spaces}. 

It should be noted that, the substask spaces are often not independent but intertwined together as the image of the common configuration space. Therefore, solving the multi-task problem requires coordination between multiple robot body parts in a complex way. 

\subsection{Optimization Problem for Velocity-based Control}
Consider a robot with its configuration space $\mathcal{C}$ given by a smooth $d$-dimensional manifold. We assume that the configuration space $\CC$ admits global coordinates, called \emph{generalized coordinates}, denoted $\q \in \mathbb{R}^d$. An example of generalized coordinates is the joint angles for a robot manipulator. In contrast to \flow, which considers acceleration policies, we are interested instead in encoding motion as a feedback velocity policy, i.e. $\qd = \pi(\q)$. Such velocity-control problem usually occurs when there is a low-level tracking controller~\cite{craig2009introduction} applied in conjunction with the policy $\pi$.

We assume that the overall task can be decomposed as a set of $K$ \emph{subtasks} defined on different \emph{subtask spaces}, denoted $\{\TT_{k}\}_{k=1}^K$. Let $\psi_k:\CC\to\TT_k$ be the subtask map for the $k$-th subtask, and let $\z_k\in\R^n$ be the generalized coordinates on the subtask space $\TT_k$, i.e., $\z_k=\psi_k(\q)$. We describe the $k$-th \emph{subtask policy} as a tuple $(\vb_k, \M_k)$, consisting of a nominal velocity policy $\vb_k:\R^n\to\R^n$ along with a state-dependent matrix-valued importance weight matrix $\M_{k}:\R^n\to\R^{n\times n}_{++}$.
The importance weight matrix $\M_{k}(\z_k)$ denotes the directional importance of the velocity policy $\vb_{k}(\z_k)$ at point $\z_k$. %

Given a collection of subtask policies $\{(\vb_k, \M_{k})\}_{k=1}^K$, our goal is to generate a {structured} configuration space velocity policy $\pi$ which trades off the error to the policies $\vb_k$ viewed on each subtask space with an importance weight defined by $\M_{k}$. Formally, the policy is given by the solution to the following weighted least-squares problem:
\begin{align}\label{eq:motion}
    &\pi(\q) \coloneqq \underset{\ub}{\argmin}\,\sum_{k=1}^K\,\big\|\,\vb_k(\psi_k(\q)) - \J_{k}(\q)\,\ub\,\big\|^2_{\Mb_k(\psi_k(\q))}
\end{align}
where $\J_{k}=\partial_\q \psi_k$ is the Jacobian of the subtask map $\psi_k$. To look deeper into the objective~\eqref{eq:motion}, the term $\vb_k(\psi_k(\q))=\vb_k(\z_k)$ is the desired velocity in the subtask space $\TT_k$, and the term $\J_k(\q)\ub$ is the velocity in the substask space $\TT_k$ if apply configuration space velocity $\ub$. Therefore, the objective function~\eqref{eq:motion} seeks to minimize the sum of deviation in each subtask space weighted by the importance weight $\M_k(\psi_k(\q))=\M_k(\z_k)$. 

\begin{figure}
\centering
\includegraphics[trim=10 0 0 10, clip, width=0.5\columnwidth]{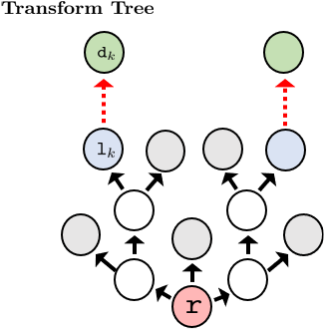}
\caption{\small{A transform tree with root in the configuration space alongside hand-specified subtask/leaf nodes (grey) and learned subtask nodes (blue). Each learned subtask node is linked to a latent subtask node (green) under a chained map $\psi_{\ltt_k\to\dtt_k} = \psi_1\circ\dots\circ\psi_M$.}}
\label{fig:tree}
\end{figure}

\subsection{The Algorithm for Policy Composition}\label{sec:alg}

Although the subtask maps $\{\psi_k\}_{k=1}^K$ can be viewed as independent when solving~\eqref{eq:motion}, the evaluation of $\{\psi_k\}_{k=1}^K$, and similarly, their Jacobians $\{J_k\}_{k=1}^K$, can benefit from \emph{reusing computation}. As an example, for robots with kinematic chain structure, the poses of the earlier links (closer to the base) are implicitly computed while evaluating the poses of the end effector. Such structure in the subtask maps can therefore lend itself amenable to computationally efficient algorithms. 

Similar to~\flow~\cite{cheng2018rmpflow}, we use a \emph{transform tree} to describe a tree-structured map from the configuration space to subtask spaces. Each node $\utt$ along the transform tree is associated with a manifold $\MM$, each edge $\ett_j$ corresponds to a smooth map $\psi_{\ett_j}\coloneqq \psi_{\vtt_j;\utt}$ from the parent node manifold to the manifold associated with child node $\vtt_j$. The root node in the transform tree, $\rtt$, corresponds to the configuration space $\CC$, and the leaf nodes $\{\ltt_k\}_{k=1}^K$ are associated with subtask spaces $\{\TT_{k}\}_{k=1}^K$. 
The subtask map is then computed as $\psi_k=\psi_{\ltt_k;\rtt}$, i.e., through aggregating the maps from the root node all the way to the leaf node $\ltt_k$. 

We propose a computational framework for solving~\eqref{eq:motion} through propagating information along the transform tree. The algorithm consists of the following four stages:
\begin{enumerate}[leftmargin=*]
    \item \textit{Forward pass:}  From the root node to the leaf nodes, the coordinate associated with each intermediate node is calculated based on the coordinate of its parent node: $\y_j = \psi_{\ett_j}(\x)$, where $\x$ and $\y_j$ are the coordinates for the parent and the child node, respectively, and $\psi_{\ett_j}$ is the map associated with the edge. The Jacobian matrix associated with each edge, $\J_{\ett_j}$, is also evaluated.  
    \item \textit{Leaf evaluation:} For each leaf node, evaluate the subtask velocity policy $\vb_k(\z_k)$ and $\Mb_k(\z_k)$. Then compute their product $\p_k(\z_k) = \M_k(\z_k)\vb_k(\z_k)$. 
    \item \textit{Backward pass:} From the leaf nodes to the root node, recursively compute the polices at each node based the policies at the child nodes: Consider a node $\utt$ with $N$ child nodes $\{\vtt_j\}_{j=1}^N$. The policy at $\utt$ is calculated as,
    \begin{align}\label{eq:back_pass}\textstyle
        \p_\utt = \sum_{j=1}^N\,\J_{\ett_j}^\t\,\p_{\vtt_j},\;\;\; \M_\utt = \sum_{j=1}^N\,\J_{\ett_j}^\t\,\M_{\vtt_j}\,\J_{\ett_j}.
    \end{align}
    where $\ett_j$ is the edge from $\utt$ to $\vtt_j$.
    \item \textit{Resolve:} At the root node, the velocity policy is solved as $\pi(\q) = \M_\rtt^{-1}\,\pb_\rtt$. 
\end{enumerate}

\subsection{Stability Properties of the Proposed Algorithm}
The configuration space motions governed by~\eqref{eq:motion} exhibit several desirable properties if the leaf node velocity policies on the transform tree take the form:
\begin{align}\label{eq:natural_gd}
    \vb_k(\z_k) = -\M_k^{-1}(\z_k)\,\nabla_{\z_k}\Phi_k(\z_k),
\end{align}
where $\Phi_k:\R^n\to\R$ is called the potential function.
We call~\eqref{eq:natural_gd} the natural gradient flow dynamics, which can be viewed as a continuous-time version of {natural gradient descent}~\cite{amari1998natural}. It evolves along steepest descent direction of $\Phi_k$ on a Riemannian manifold defined by Riemannian metric $\M_k$. Under the assumption that each leaf node policy is given by a natural gradient descent system~\eqref{eq:natural_gd}, the following properties hold for the generated root node velocity policy:
\begin{flushitemize}
    \item \emph{Closure: }the motion follows natural gradient flow with metric $\M_\rtt=\sum_{k=1}^K \J_k^\t \M_k\J_k$, and potential function $\Phi_\rtt=\sum_{k=1}^K \Phi_{k}\circ\psi_{k},$
    where $\circ$ denotes map composition;
    \item \emph{Stability: }the system converges to the stationary points of the potential function $\Phi_\rtt$. 
     
\end{flushitemize}

Formally, the above properties are stated in the following theorem:
\begin{theorem}\label{thm:stability}
Assume that the importance weight matrix at the root node is non-singular, i.e. $\Mb_\rtt\succ0$. If each subtask policy is given by natural gradient flow~\eqref{eq:natural_gd}, the root node policy is given by natural gradient flow $\qd = -\M_\rtt^{-1}\nabla_\q\Phi_{\rtt}$, where $\Phi_\rtt=\sum_{k=1}^K \Phi_{k}\circ\psi_{k}$. Further, if $\Phi_\rtt$ is proper, continuously differentiable and lower bounded, the system $\qd=\pi(\q)$ converges to a forward invariant set $\CC_\infty \coloneqq \{\q : \nabla_\q \Phi_\rtt = 0\}$.
\end{theorem}
\begin{proofsketch}
Assume each subtask policy is given by natural gradient flow, $\pb_{k}  = \M_{k}\,\vb_{k} = -\nabla_{\z_k}\Phi_{k}$, for all $k\in\{1,\ldots,K\}$. We will prove that each node follows natural gradient flow: Consider any non-leaf node $\utt$. Let $\{\vtt_j\}_{j=1}^N$ be the child nodes of $\utt$. Suppose each child node $\vtt_j$ follows natural gradient flow with potential $\Phi_{\vtt_j}$ and metric $\M_{\vtt_j}$. At node $\utt$, by~\eqref{eq:back_pass},
\begin{align}\textstyle
    \p_\utt = \sum_{j=1}^N\,\J_{\ett_j}^\t\,\p_{\vtt_j} = \sum_{j=1}^N\,\J_{\ett_j}^\t\,\nabla_{\y_j}\Phi_{\vtt_j} = \nabla_\x\Phi_\utt,\\
    \nonumber\text{where }\Phi_\utt \coloneqq  \sum_{j=1}^N\Phi_{\vtt_j} \circ \psi_{\ett_j}
\end{align}
Therefore, by recursively applying the analysis from the leaf nodes to the root node, we have that the root node also follows natural gradient flow $\pb_\rtt = \nabla_\q\Phi_\rtt$. Hence, we have,
\begin{equation}\textstyle
    \begin{split}
        \frac{d}{dt}\,\Phi_\rtt &= \qd^\t\,\nabla_\q\Phi_\rtt = -\big(\nabla_\q\Phi_\rtt\big)^\t \M_\rtt^{-1}\,\nabla_\q\Phi_\rtt
    \end{split}
\end{equation}
Under the assumption $\M_\rtt\succ 0$, by LaSalle's invariance principle~\cite{khalil2002nonlinear}, the system converges to the forward invariant set $\CC_\infty = \{\q : \nabla_\q \Phi_\rtt = 0\}$.
\end{proofsketch}

\vspace{2mm}
\textbf{The Benefit of Velocity-based Motion Control: }The main benefit of our proposed velocity-based motion control framework, compared to \flow~\cite{cheng2018rmpflow}, is its simplicity. The forward pass and backward pass of~\flow requires computing additional curvature terms $\Jd_k\qd$ resulting from pushing forward accelerations. More importantly, to ensure the stability of the generated motion, it also requires the leaf policies to take a more complicated form, known as geometrical dynamical systems~\cite{cheng2018rmpflow}, which involves curvature terms of the importance weight matrices $\{\M_k\}_{k=1}^K$. This creates a huge amount of computational overhead for learning when the importance weight matrices are parameterized, and also makes the optimization problem more complicated. 


\section{Learning Structured Motion Policies for Human Demonstrations}\label{sec:approach}
In this section, we provide details of our approach to learning motion policies~\eqref{eq:motion} from human demonstrations. 

\subsection{Problem Statement}\label{sec:problem_statement}

Consider the problem of kinesthetic teaching, where the human provide demonstrations by moving the robot, providing a number of trajectory demonstrations in the joint configuration space. Additionally, we allow the human to specify a number of \emph{subtask spaces}\footnote{In most existing learning from demonstrations literature\cite{paraschos2017probabilistic, pastor2009learning,khansari2011learning,neumann2015learning,chaandar2019learning, rana2020euclideanizing}, only a single (subtask) space is considered, and it is usually either the configuration space, or the (3-d or 6-d) end-effector workspace. }, where the motion on the these subtask spaces is relevant to the achieving the overall task. For example, for a goal-reaching subtask, the user may specify, as a subtask space, the 3-dimensional Euclidean workspace of the end-effector position. The goal for our learning problem is to learn a parameterized policy $\pi^\theta$ which can generate motion similar to the human demonstrations when \textit{viewed in these subtask spaces}. 

Consider $N$ trajectory demonstrations in the configuration space of the robot, each composed of $T_i$ position-velocity pairs, denoted by $\{\{({\q}_{i,t}^d, \dot{{\q}}^d_{i,t})\}_{t=1}^{T_i}\}_{i=1}^{N}$. On possible way of learning  is to directly regress joint space velocity so that it matches the joint velocity of the demonstrations, i.e.,
\begin{equation}\label{eq:joint_loss}
    \theta^\star_\CC = \underset{\theta}{\arg\min}\,  \underbrace{\sum_{i=1}^N\sum_{t=1}^{T_i}\, \bigg\|\dot{\q}_{i,t}^d - \pi^{\theta}\big({\q}^d_{i,t}\big) \bigg\|^2}_{\LL_\CC(\theta)}.
\end{equation}
Presumably, if the learned joint velocity policy perfectly matches the demonstrated joint velocities, it should also perfectly matches the demonstrations in the subtask spaces. However, in practice, there usually does not exist a policy such that $\LL_\CC(\theta)=0$. 
This is because, when providing demonstrations, the human primarily cares about the motion on spaces that are directly relevant to achieving the task (i.e. the subtask spaces). As a result, the demonstrations can be conflicting in the joint space (providing vastly different velocities at the same joint postion), even though they can be consistent in the subtask spaces. Therefore, directly regressing on the joint space trajectories~\eqref{eq:joint_loss} will produce large error both in the joint space and in the workspace.

Given the observation that the trajectory demonstrations are the most informative when considered in the subtask spaces, we proposed an alternative objective function that direct penalize the deviation in the subtask spaces:
\begin{equation}\label{eq:learning_problem}
\begin{split}
    \theta^\star &= \underset{\theta}{\arg\min}\,  \sum_{i=1}^N\sum_{t=1}^{T_i}\sum_{k=1}^{{K}}\, 
    \lambda_k\,\big\|\J_k\dot{{\q}}^d_{i,t} - \J_k\pi^\theta({\q}^d_{i,t}) \big\|^2\\
    &=\underset{\theta}{\arg\min}\,  \underbrace{\sum_{i=1}^N\sum_{t=1}^{T_i}\sum_{k=1}^{{K}}\, \lambda_k\,\big\|\dot{{\q}}^d_{i,t} - \pi^\theta({\q}^d_{i,t}) \big\|^2_{ \J_k^\t \J_k}}_{\LL(\theta)},
\end{split}
\end{equation}
where $\lambda_k>0$ is the user-specified weight for the $k$-th subtask. In contrast to~\eqref{eq:joint_loss}, our proposed objective only penalizes the deviation of velocities in the subtask spaces. The velocity in the \textit{subtask spaces} is given by the pushforward operator $\q\mapsto\J_k(\q)$. 

Let us now consider policy learning with the structured policy class introduced in Section~\ref{sec:background}. Conviently, as the subtask spaces is provided by the demonstrator, we can represent the joint space velocity as the solution to the motion generation problem described in Section~\ref{sec:background}, where the subtask policies are parameterized, i.e.,
\begin{align}\label{eq:motion_param}
    &\pi^\theta(\q) =\underset{\ub}{\argmin}\,\sum_{k=1}^K\,\big\|\,\vb_k^{\theta_k}(\psi_k(\q)) - \J_{k}\,\ub\,\big\|^2_{\Mb_k^{\theta_k}}.
\end{align}

We can then optimize for the objective function~\eqref{eq:learning_problem} through, e.g. gradient descent:
\begin{equation}\label{eq:update}
\begin{split}
    \theta &\gets \theta - \alpha \nabla_\theta \LL(\CC)\\
    &\gets\theta - \alpha \sum_{i=1}^N\sum_{t=1}^{T_i}\frac{\partial \pi^\theta(\q_{i,t}^d)}{\partial\theta} \nabla_{\pi^\theta(\q_{i,t}^d)}\LL(\theta),
\end{split}
\end{equation}
where the term $\frac{\partial \pi^\theta(\q_{i,t}^d)}{\partial\theta}$ can be computed by back-propagating through the motion generation algorithm descibed in Section~\ref{sec:alg}.

Note that the policy parameterization $\pi^\theta$ in~\eqref{eq:motion_param} also allows for certain subtask policies to be \emph{fixed} during learning, i.e., $\theta_k=\emptyset$. This provides the user with the freedom to manually design subtask policies, for, e.g., joint damping, respecting joint limit, collision avoidance, etc. In this case, the parameterized subtask policies also learn to trade off against the hand-designed policies through the importance weight matricies. For the remainder of this section, we present an expressive class of stable learnable subtask policies which result in a stable configuration space velocity policy under Theorem~\ref{thm:stability}.

\begin{figure}
	\centering
	\includegraphics[width=0.8\columnwidth]{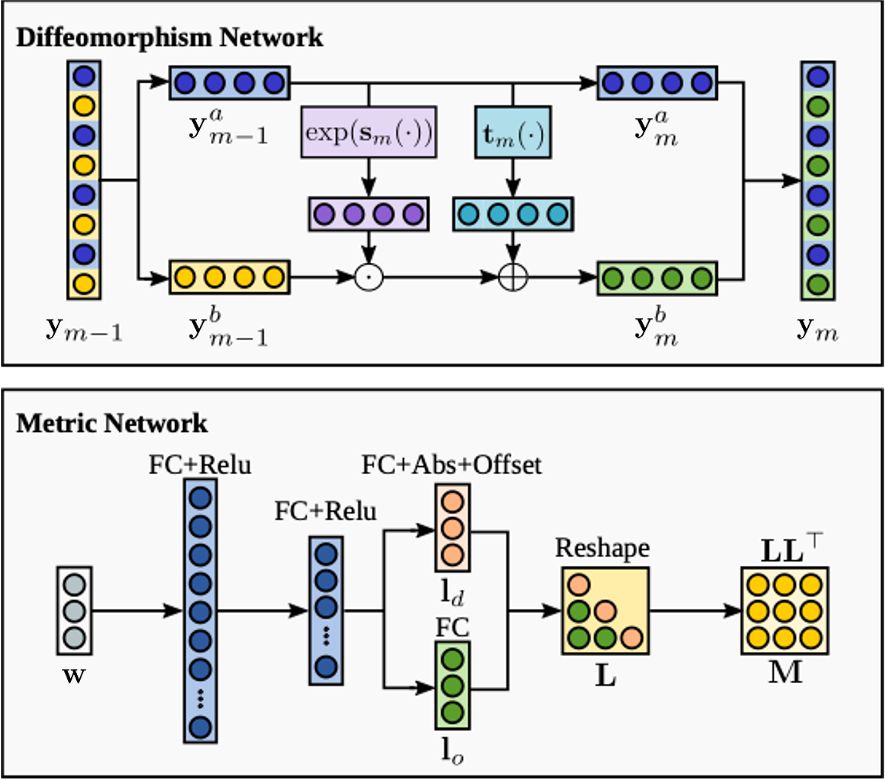}
	\vspace{-0.2cm}
	\caption{\small{\emph{Top:} Structure of the network defining a single map $\psi_m$ in the diffeomorphism chain~\cite{rana2020euclideanizing}. \emph{Bottom:} Structure of the network for defining latent subtask metric $\M_{\dtt_k}$~\cite{rana2020learning}. }}
	\label{fig:network}
\end{figure}

\subsection{A Class of Stable Subtask Policies}

We seek to paramterize subtask policies $\{(\vb_k^{\theta_k},\M_{k}^{\theta_k})\}_{k=1}^K$ so that the resulting configuration space policy is stable. According to Theorem~\ref{thm:stability}, the resulting motion is stable as long as the subtask polcies follows natural gradient flow~\eqref{eq:natural_gd}. Therefore, we choose to parameterize the subtask policy $(\vb_k,\M_{k})$ through the tuple $(\Phi_{k}^{\theta_k},\M_{k}^{\theta_k})$, where $\Phi_k$ is the potential function. The velocity policy is then given by $\vb_k^{\theta_k}=(\M_k^{\theta_k})^{-1}\nabla\Phi_k^{\theta_k}$. To ensure stability of the combined policy in configuration space, we need the importance weight matrix $\M_{k}^{\theta_k}$ to be always positive definite. Additionally, we require $\Phi_{k}$ to have a unique mimima at a desired goal location $\z_k^*$ as we care primarily about goal-directed motions. 

\subsubsection{From subtasks to latent subtasks}
The main challenge for representing the subtask policy is finding a expressive parameterization of the potential function without introducing spurious attractive points. While direct parameterization with such property is in general challenging, recent work~\cite{rana2020euclideanizing} has demonstrate success through parameterizing diffeomorphisms to \emph{latent} spaces, where a simple potential function is defined there. We adopt this approach as it shows expressiveness in representing complex motions without introducing undesired local minima. 

Conveniently, in our transform tree formulation (Section~\ref{sec:alg}), this is equivalent to adding a child node, $\dtt_k$, to the original ``leaf'' node $\ltt_k$ (see Fig.\ref{fig:tree}). The map between $\dtt_k$ and $\ltt_k$ is a learnable map, $\phi^{\theta_k}_k:\z_k\mapsto \wb_k$.

Then, in the latent space, we can use a simple pre-specified potential function, e.g. $\Phi_{\dtt_k}(\wb_k) = 0.5\|\wb_k - \phi^{\theta_k}_k(\z_k^*)\|^2$, and any positive-definite parameterization of the importance weight matrix\footnote{One can also choose to parameterize the matrix in the subtask space instead of the latent space. } $\M_{\dtt_k}^{\theta^k}$. Then, by properties of diffeomorphisms~\cite{rana2020euclideanizing}, the generated motion is guaranteed stable. 
Next, we introduce the parameterizations we choose for the diffeomorphism and importance weight matrix, respectively. 

\subsubsection{A chain of diffeomorphisms}
To realize a diffeomorphism, we rely on the formulation in~\cite{rana2020euclideanizing} (see Fig.\ref{fig:network}). Specifically, we view $\phi^{\theta_k}_k$ as a chain of $M$ simpler maps, i.e.  $\phi^{\theta_k}_k = \psi_{1}\circ\dots\circ\psi_{M}$. Assuming coordinates $\y_{m}\in\R^n$ for the co-domain of $\psi_{m}$ i.e. $\y_{m}=\psi_m(\y_{m-1})$, $\y_0=\z_k$, and $\y_M=\wb_k$, we define, 
\begin{equation}\label{eq:coupling_layer}\textstyle
    \y_m = 
    \begin{bmatrix}
    \y_m^a\\\y_m^b
    \end{bmatrix} = 
    \begin{bmatrix}
    \y_{m-1}^a\\
    \y_{m-1}^b \odot \exp \big( {s}_m(\y_{m-1}^{a}) \big) + {t}_m(\y_{m-1}^{a})
    \end{bmatrix}, 
\end{equation}
where $\odot$ and $\exp$ denote pointwise product and pointwise exponential respectively. The components $\y_{m-1}^a\in\R^{\lfloor n/2\rfloor}$ and $\y_{m-1}^b\in\R^{\lceil n/2 \rceil}$ constitute alternate input dimensions, with the pattern of alternation reversed after each mapping in the chain. Furthermore,  ${s}_m:\R^{\lfloor n/2\rfloor}\to\R^{\lceil n/2\rceil}$ and ${t}_m:\R^{\lfloor n/2\rfloor}\to\R^{\lceil n/2\rceil }$ are learnable scaling and translation functions, respectively. We parameterize the scaling and translation functions as linear combinations of random Fourier features~\eqref{eq:coupling_layer} i.e. $s_m(\cdot) \coloneqq s_m(\cdot\,; \theta_{s_m}) = \varphi(\cdot)^\t\theta_{{s}_m}$, and $t_m(\cdot) \coloneqq t_m(\cdot\,; \theta_{t_m}) = \varphi(\cdot)^\t\theta_{{t}_m}$, where 
\begin{equation}\label{eq:rff}\textstyle
    \varphi(\cdot) = \sqrt{\frac{2}{D}}\big[\cos(\bm{\alpha}_1^\t(\cdot) + \bm{\beta}_1), \dots, \cos(\bm{\alpha}_D^\t(\cdot) + \bm{\beta}_D)\big]^\t \otimes \mathbf{I}, 
\end{equation} 
is a $D$-dimensional Fourier feature approximation of a matrix-valued Gaussian separable kernel \cite{alvarez2011kernels,sindhwani2012scalable}, $K(\y,\y') = \exp (-\frac{\|\y-\y'\|^2}{2l^{2}}) \mathbf{I}$ with length-scale $l$. Due to the choice of parameterization in \eqref{eq:coupling_layer}-\eqref{eq:rff}, $\psi_m$ is a smooth and affine bijective map, and thus a diffeomorphism. Consquently, the chain $\phi^{\theta_k}_k$ is a diffeormorphism. 

\subsubsection{Importance weight matrix via Cholesky decomposition}
Similar to~\cite{rana2020learning}, we represent a latent subtask inertia matrix $\M_{\dtt_k}$ by its Cholesky decomposition parameterized by a matrix-valued neural network (see Fig.\ref{fig:network}). This parameterization has been previously introduced in literature~\cite{rana2020learning}. Concretely, we construct $\M_{\dtt_k} \coloneqq \Lb_{\dtt_k}\Lb_{\dtt_k}^\t$, where $\Lb_{\dtt_k}(\wb_k) \in \mathbb{R}^{n\times n}$ is a lower-triangular matrix. We parameterize the vectorized diagonal and off-diagonal entries of $\Lb_{\dtt_k}$, i.e. $\lb_d(\wb_k; \theta_{\lb_d})\in\R^n$ and $\lb_o(\wb_k; \theta_{\lb_o})\in\R^{\frac12 (n^2 - n)}$ respectively, as fully-connected neural networks with RELU activations. Furthermore, the networks for $\lb_o$ and $\lb_d$ share parameters for all the layers except their output layers. To ensure $\Lb_{\dtt_k}$ is a valid Cholesky decomposition, and consequently $\Mb_{\dtt_k}$ is positive definite, we require the entries of $\lb_d$ to be strictly positive.  In lieu of this, we take the absolute value of the output linear layer of $\lb_d$ and add a small positive bias $\epsilon>0$. %

\subsection{Discussion}
At first glance, our approach may seems very similar to~\cite{rana2020learning} as we both learn multiple subtask policies and, during execution, combine them together. However, mathematically, they are fundamentally different. 

In~\cite{rana2020learning}, each subtask policy is independently learned to imitate the demonstrated trajectories mapped the the corresponding space, i.e.,
\begin{equation}\label{eq:learning_problem_old}
\begin{split}
    \theta^\star_k &= \underset{\theta_k}{\arg\min}\,  \sum_{i=1}^N\sum_{t=1}^{T_i}\,\big\|\J_k\dot{{\q}}^d_{i,t} - (\M_k^{\theta_k})^{-1} \nabla\Phi_k^{\theta_k}\big\|^2.
\end{split}
\end{equation}

The combination of these individually-learned policy only happens after learning, i.e., during the execution of the policy. This strategy has the following limitations. First, the learned importance weight matrices are not learned to provide the proper trade-offs between policies, as each policy is learned independently. Due to this, additional manual scaling of the importance weight matrices is needed especially when combined with hand-designed policies. Second, the geometric constraints (e.g. induced by the kinematics of the robot) between tasks are not considered during learning, which contributions to error during execution. 
In summary, despite its empirical success, the approach introduced in~\cite{rana2020learning} is not able to fully exploit the benefits provided by the policy structure. Whereas our approach, by properly formulating the learning problem~\eqref{eq:learning_problem} and differentiable through the structured policy~\eqref{eq:motion_param}, is able to take full advantage over the policy structure during learning. We will further demonstrate this through experiments in upcoming section. 


\section{Experimental Results}\label{sec:results}
\begin{figure}[htb!]
   \centering
   	\begin{subfigure}{1\linewidth}
 	    \includegraphics[trim=270 0 0 50, clip, width=0.24\linewidth]{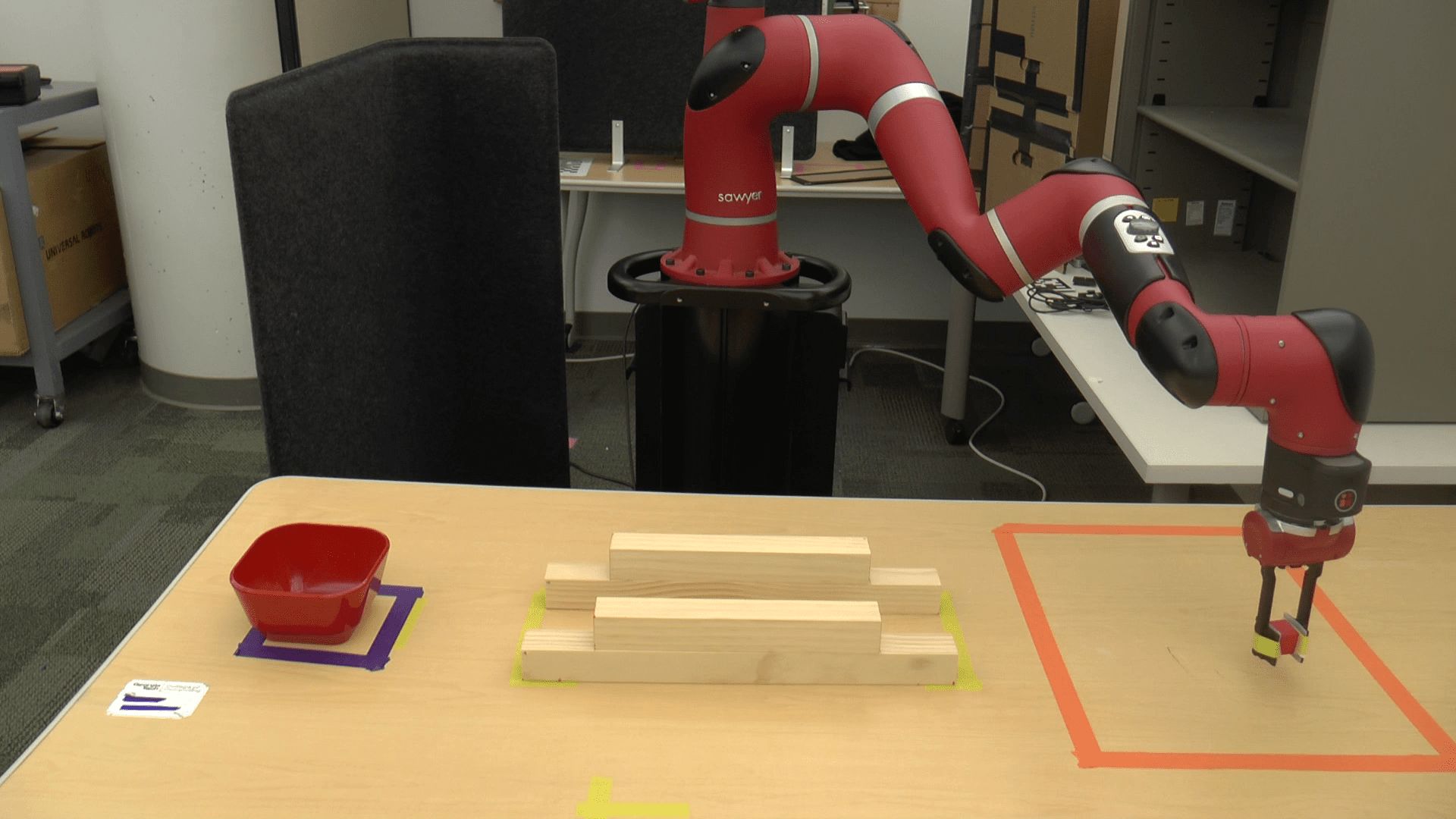}
 	    \includegraphics[trim=270 0 0 50, clip, width=0.24\linewidth]{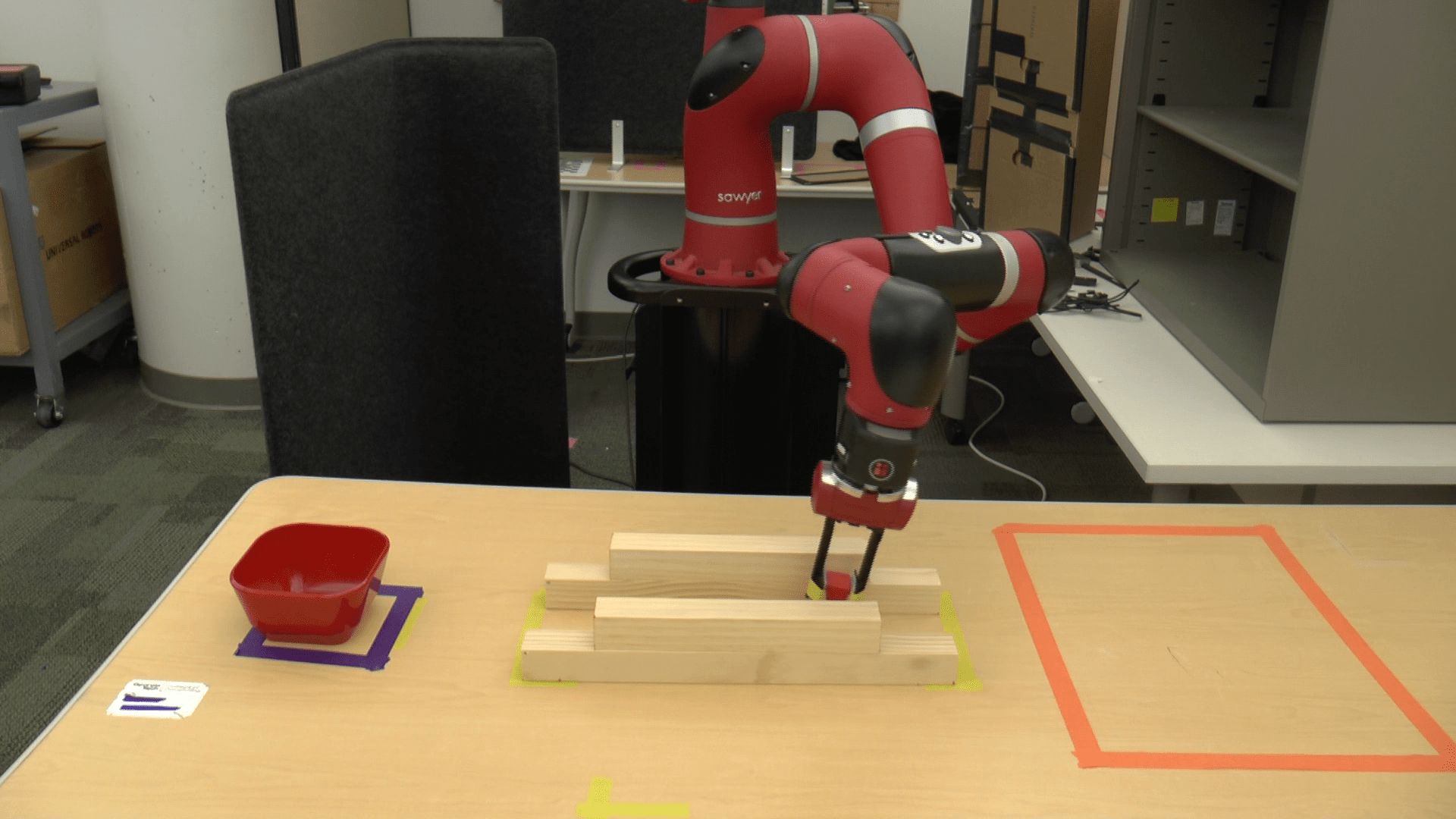}
 	    \includegraphics[trim=270 0 0 50, clip, width=0.24\linewidth]{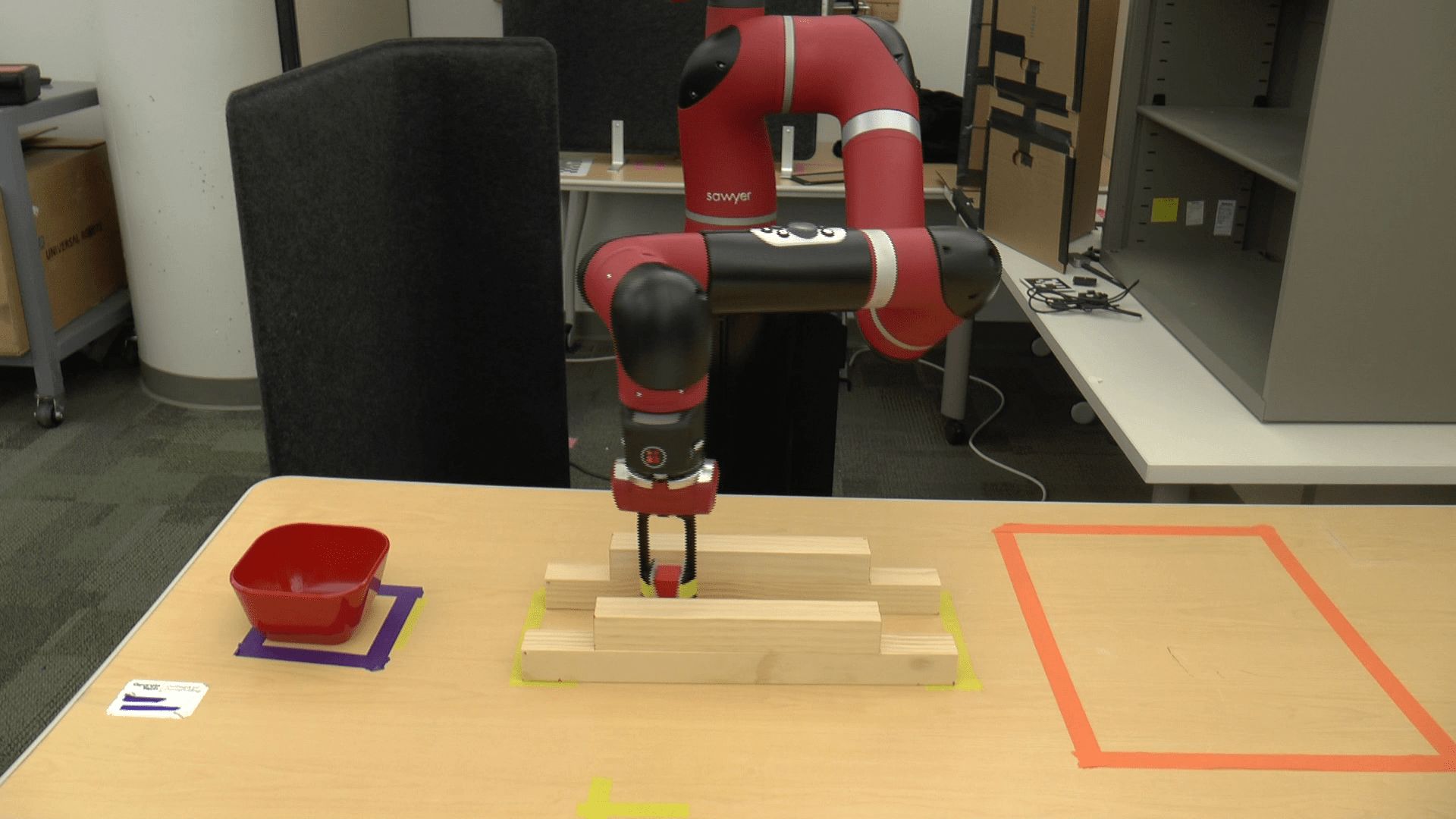}
 	    \includegraphics[trim=270 0 0 50, clip, width=0.24\linewidth]{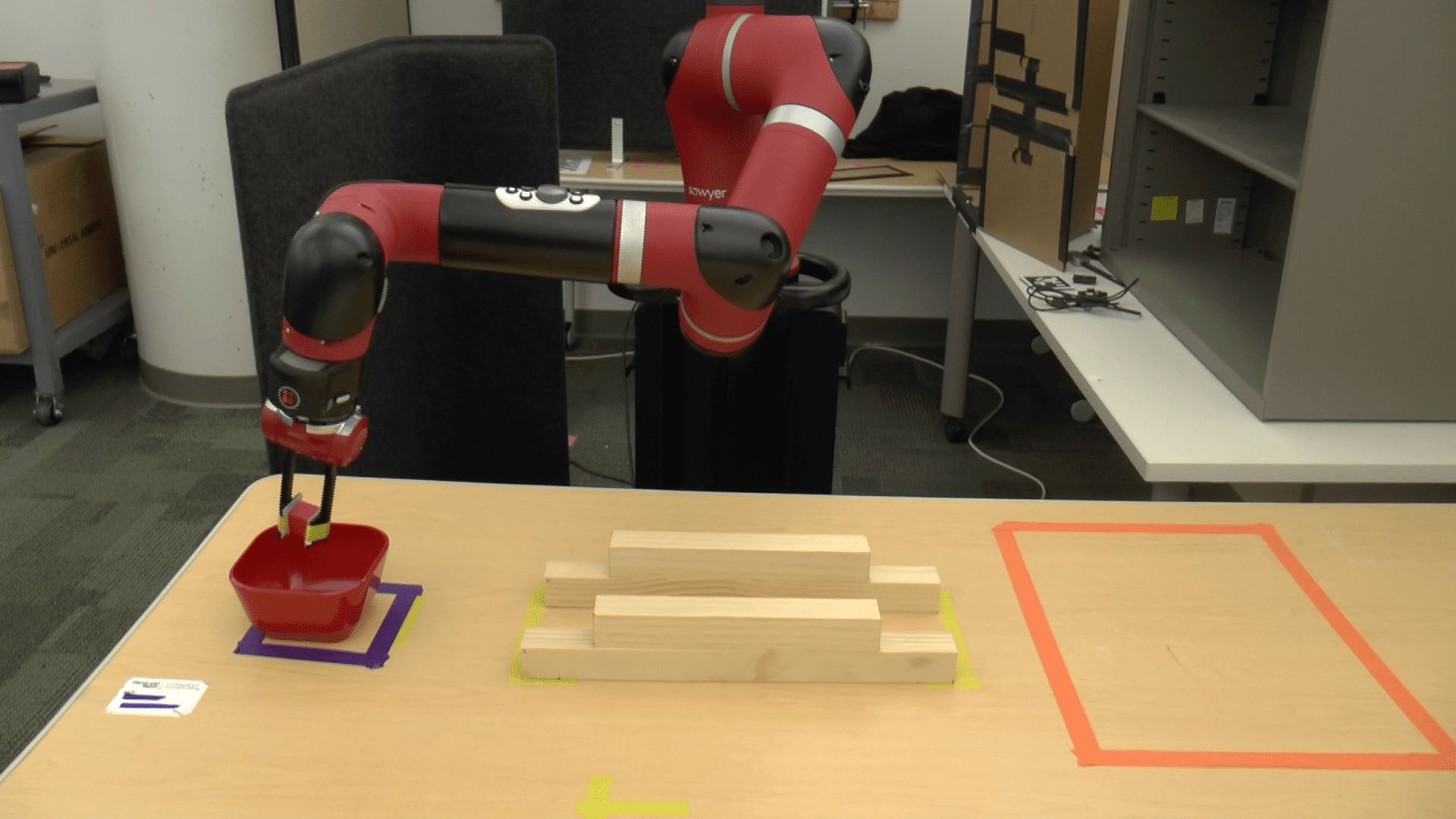}
   	\end{subfigure}
   	 \begin{subfigure}{1.0\linewidth}	
	 	\includegraphics[trim=0 0 0 0, clip, width=\linewidth]{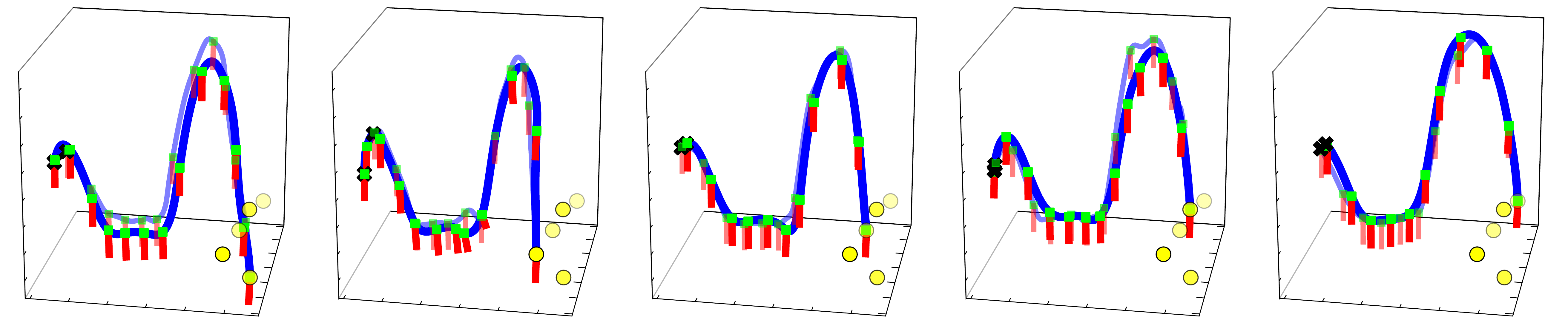}
   	\end{subfigure}
\caption{\small{The \emph{inspection} task required the robot to pick an object from one side of the table and place it in a bowl on the other side. In the middle, the robot was required to pass a constrained pathway. \emph{Top:} A series of snapshots showing a robot executing learned behavior. \emph{Bottom:} Plots of a subset of motion reproductions from different initial poses, overlayed on corresponding demonstrations. The yellow circles represent the initial end-effector positions, each corresponding to one of the rollouts.}}
\label{fig:inspection}
\end{figure}

\begin{figure}[htb!]
   \centering
   	\begin{subfigure}{1\linewidth}
		\includegraphics[trim=390 0 480 130, clip, width=0.24\linewidth, height=2cm]{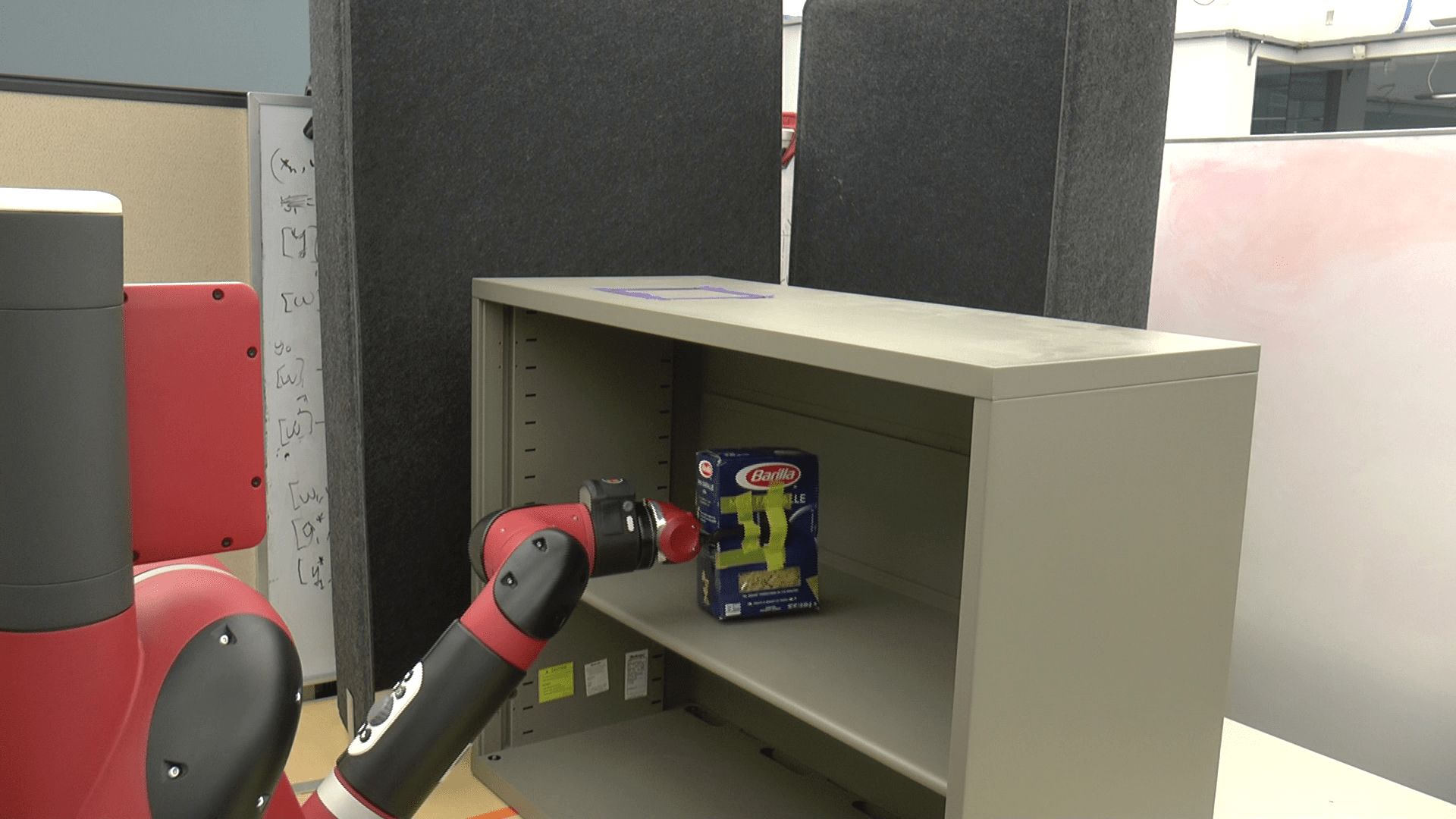}
		\includegraphics[trim=390 0 480 130, clip, width=0.24\linewidth, height=2cm]{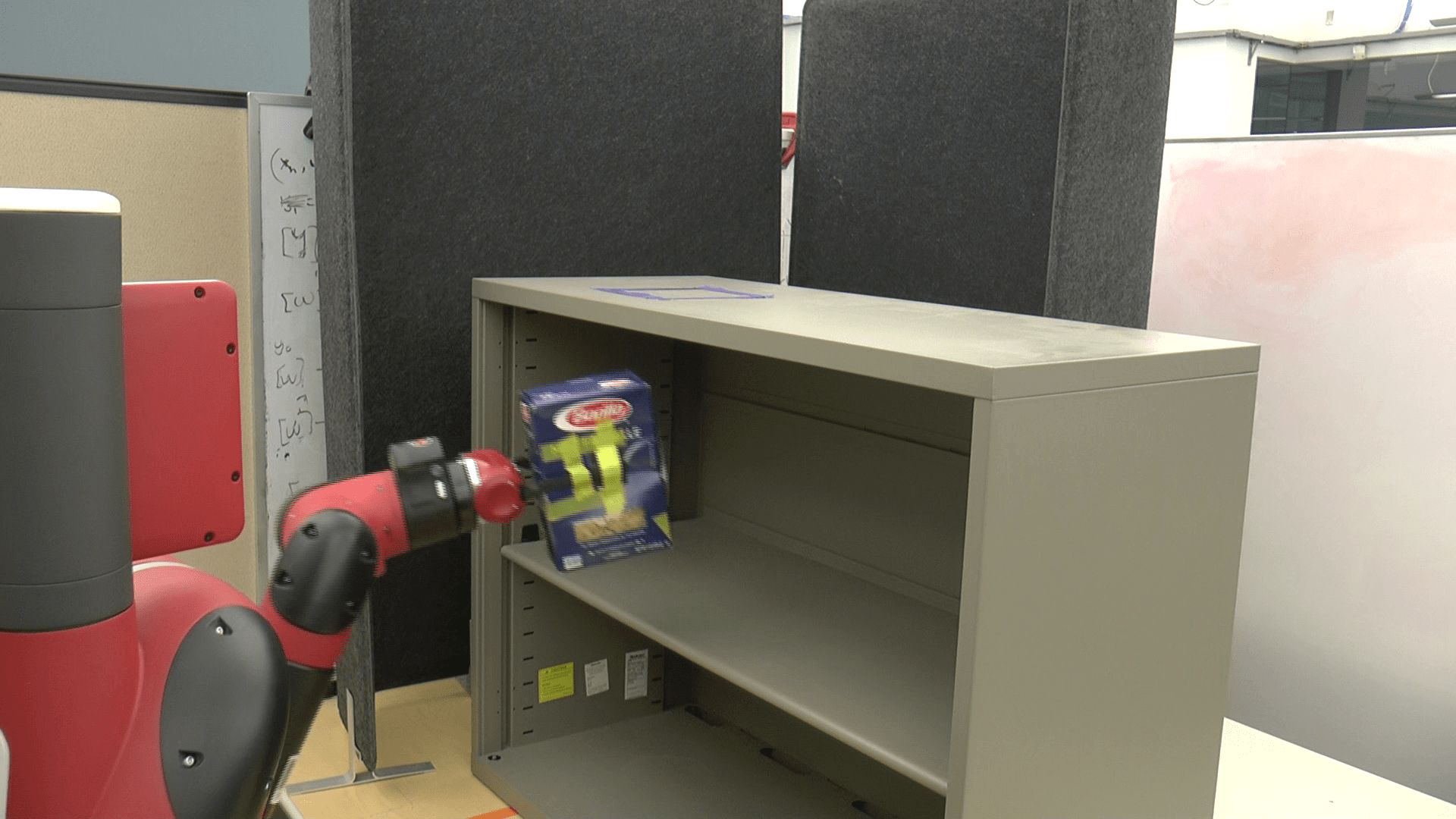}
		\includegraphics[trim=390 0 480 130, clip, width=0.24\linewidth, height=2cm]{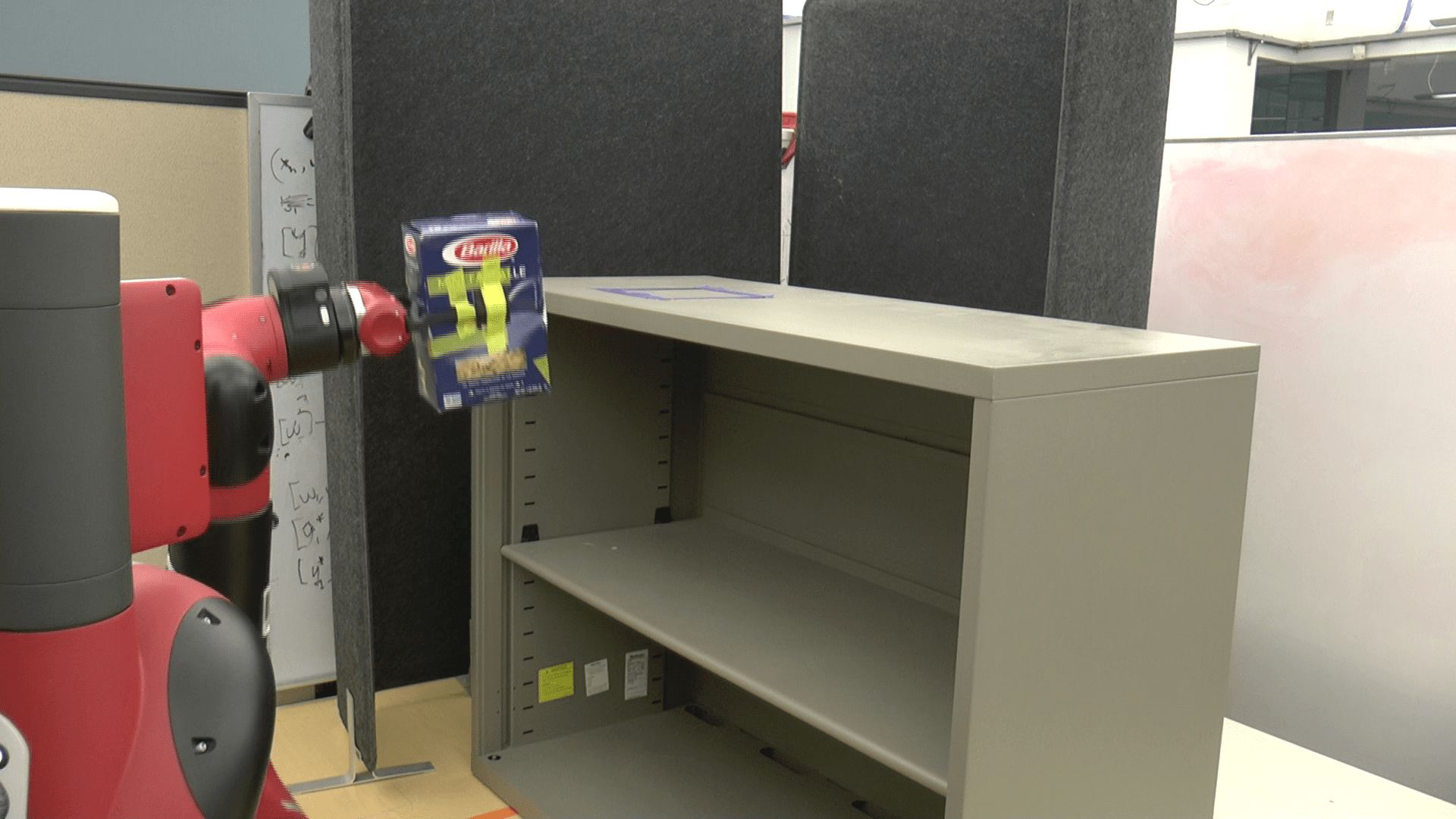}
		\includegraphics[trim=390 0 480 130, clip, width=0.24\linewidth, height=2cm]{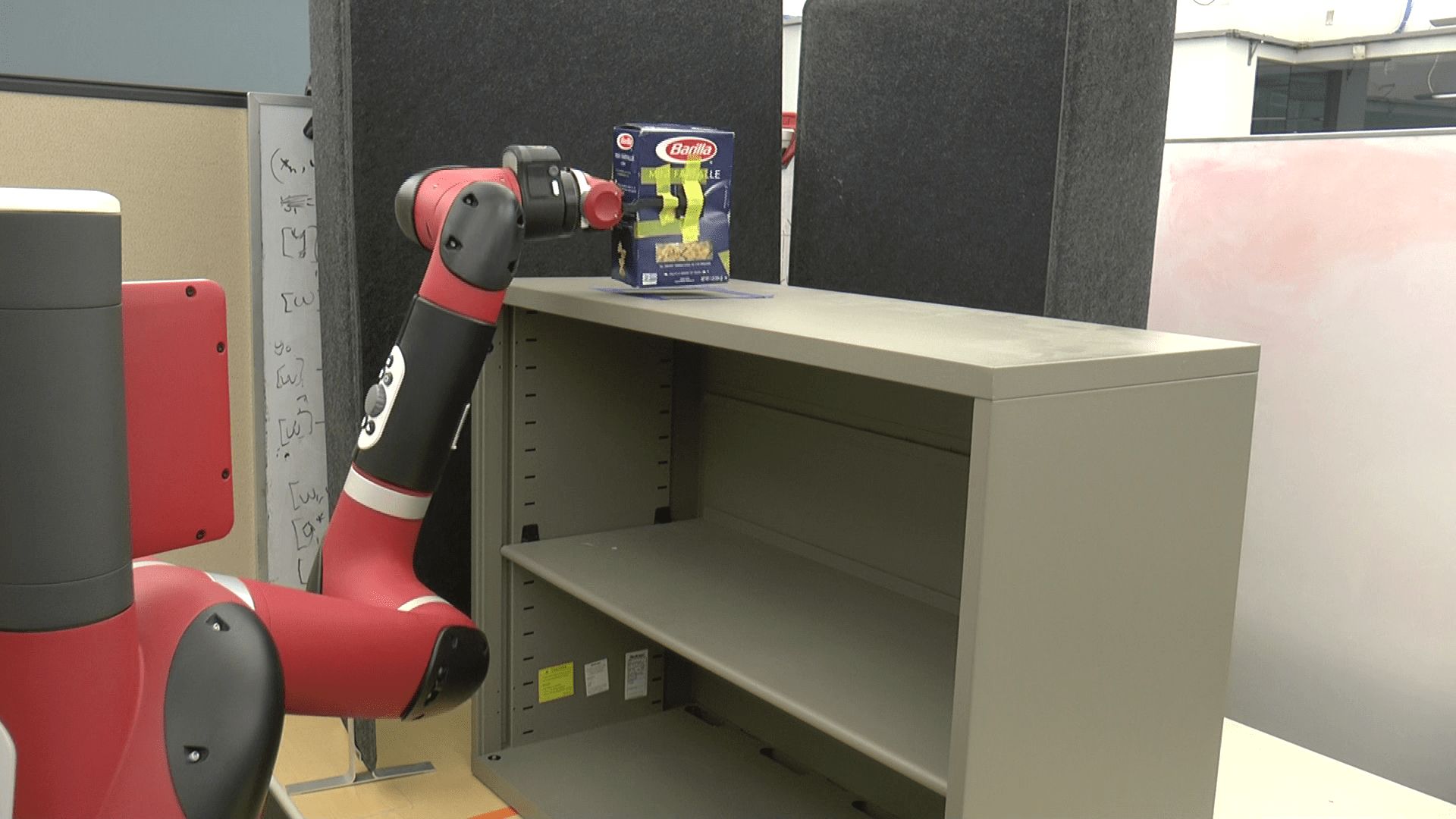}
   	\end{subfigure}
   	 \begin{subfigure}{1.\linewidth}	
	 	\scalebox{-1}[1]{\includegraphics[trim=0 0 0 0, clip, width=\linewidth]{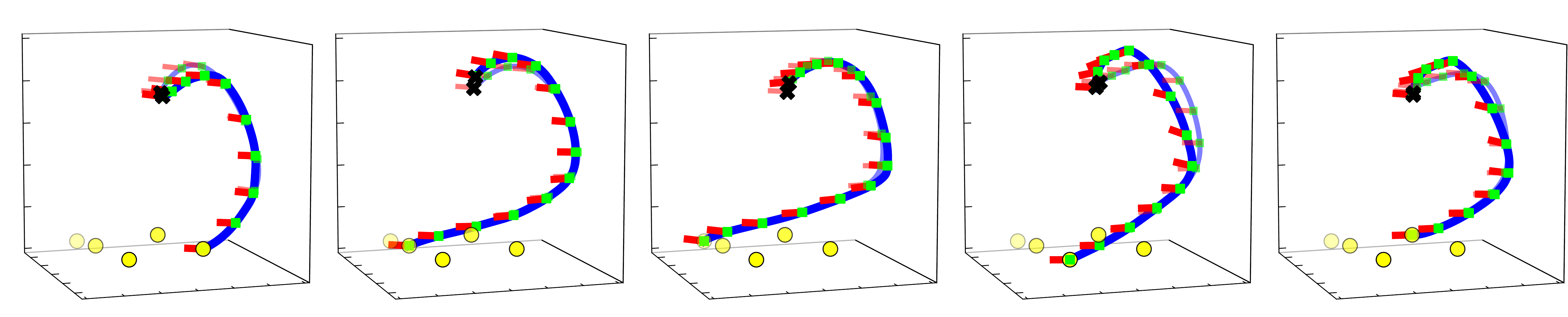}}
   	\end{subfigure}
\caption{\small{The \emph{placing-1} task required the robot pick an object from a lower shelf and place it on at a goal location on the top-most shelf at a certain orientation. \emph{Top:} A series of snapshots showing a robot executing learned behavior. \emph{Bottom:} Plots of a subset of motion reproductions from different initial poses, overlaid on corresponding demonstrations. The yellow circles represent the initial end-effector positions, each corresponding to one of the rollouts.}}
\label{fig:placing_1}
\end{figure}

\begin{figure}[htb!]
   \centering
   	\begin{subfigure}{1\linewidth}
 	    \includegraphics[trim=400 0 450 0, clip, width=0.24\linewidth, height=2cm]{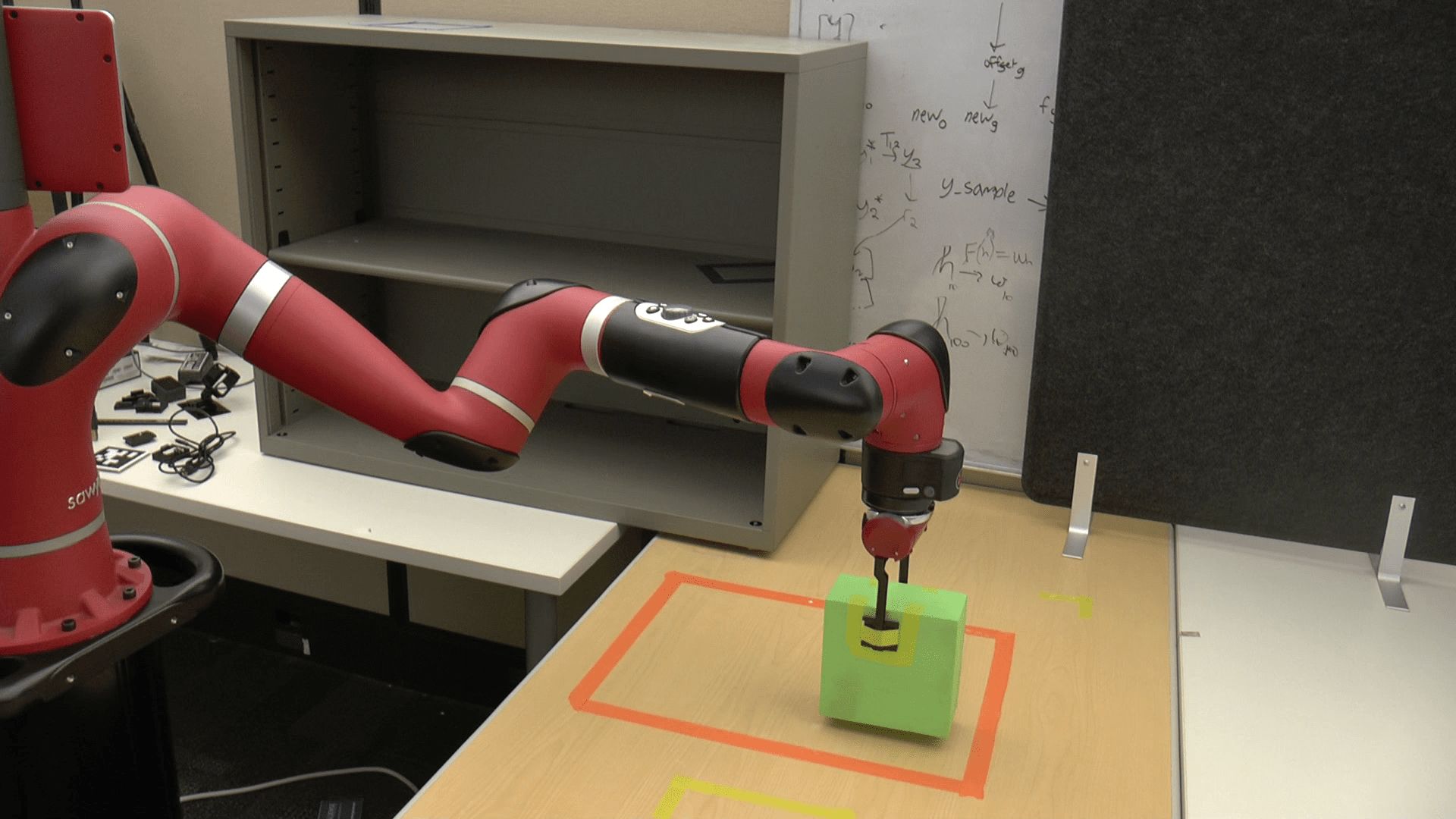}
 	    \includegraphics[trim=400 0 450 0, clip, width=0.24\linewidth,height=2cm]{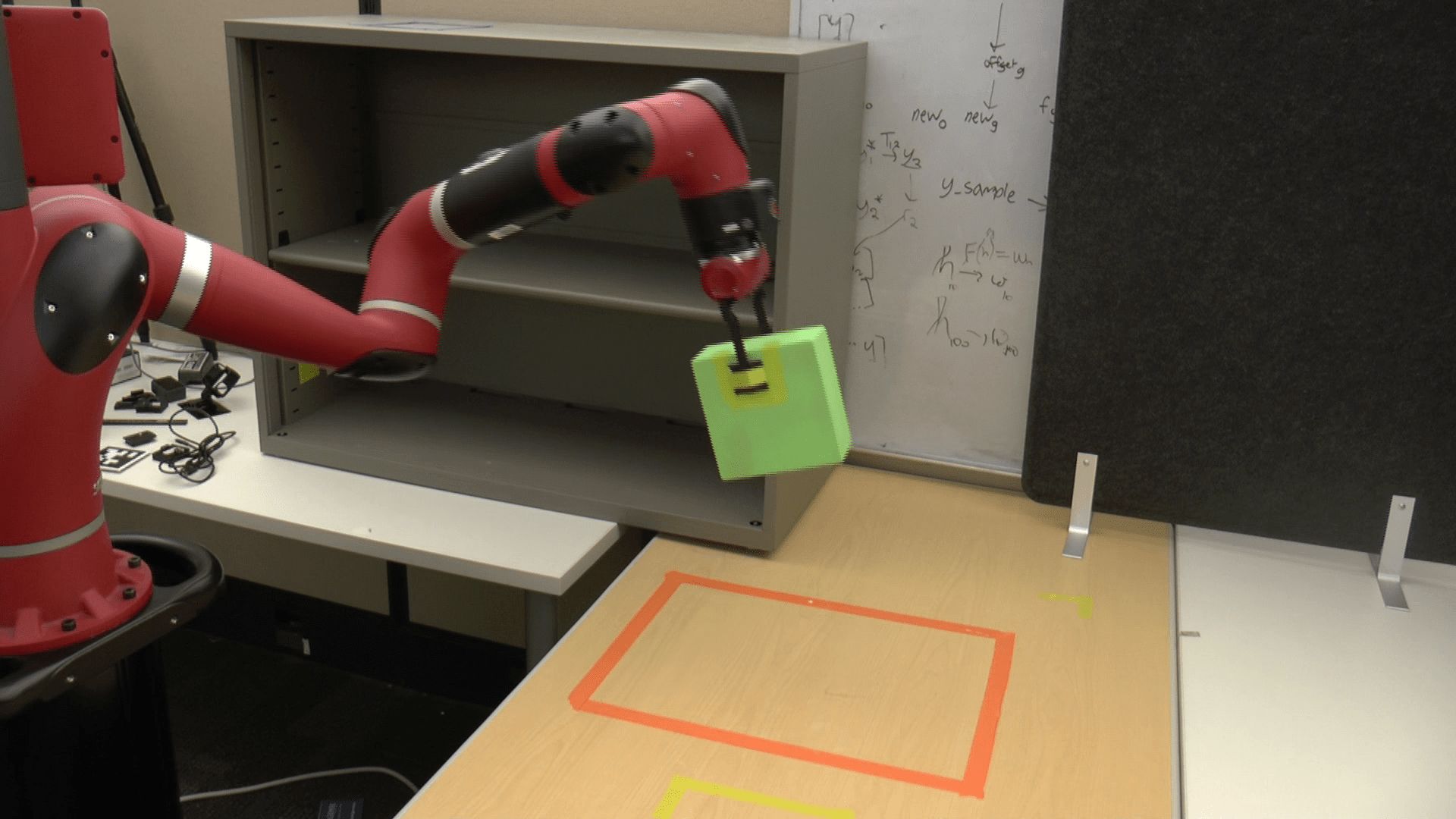}
 	    \includegraphics[trim=400 0 450 0, clip, width=0.24\linewidth,height=2cm]{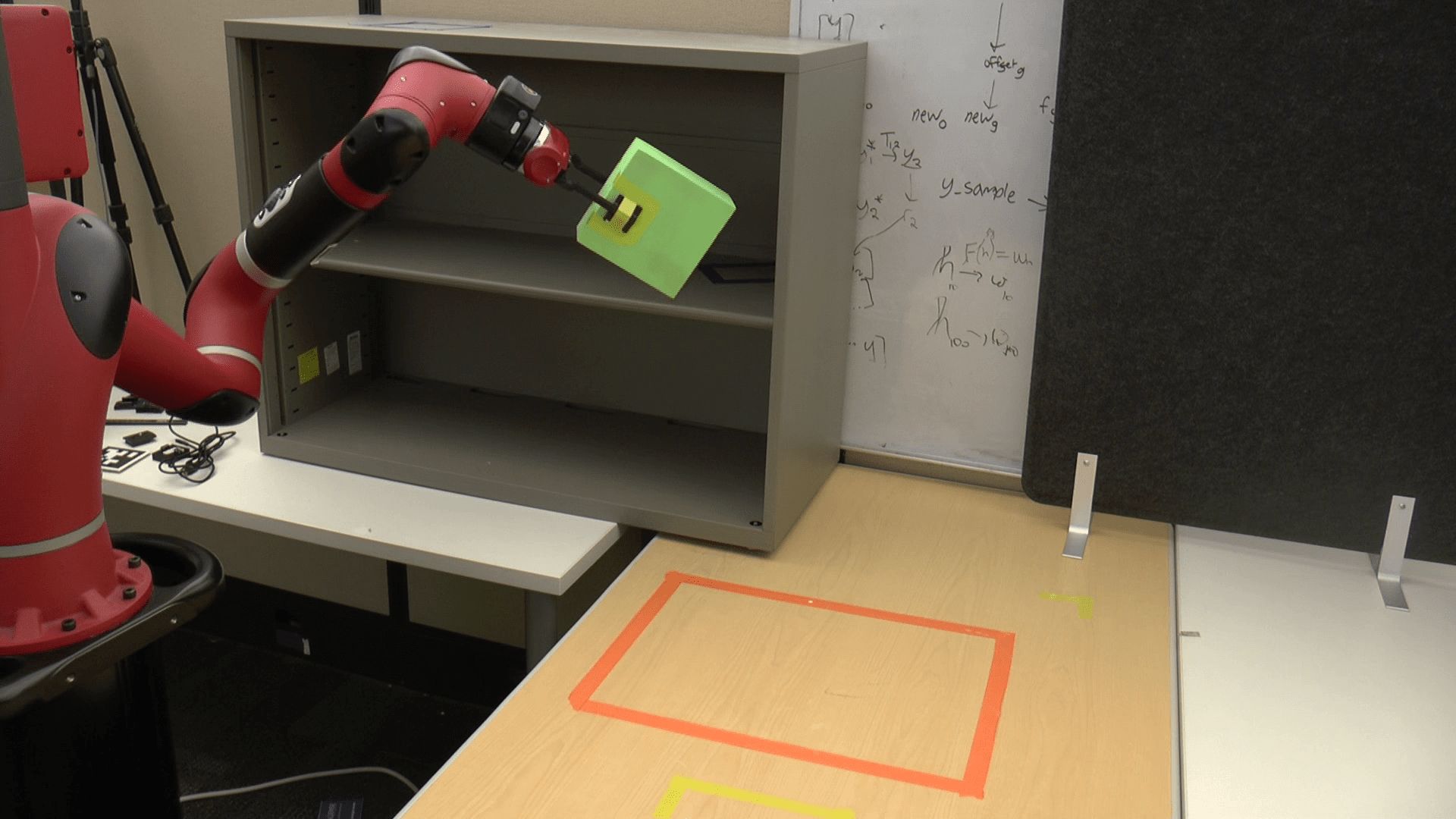}
 	    \includegraphics[trim=400 0 450 0, clip, width=0.24\linewidth,height=2cm]{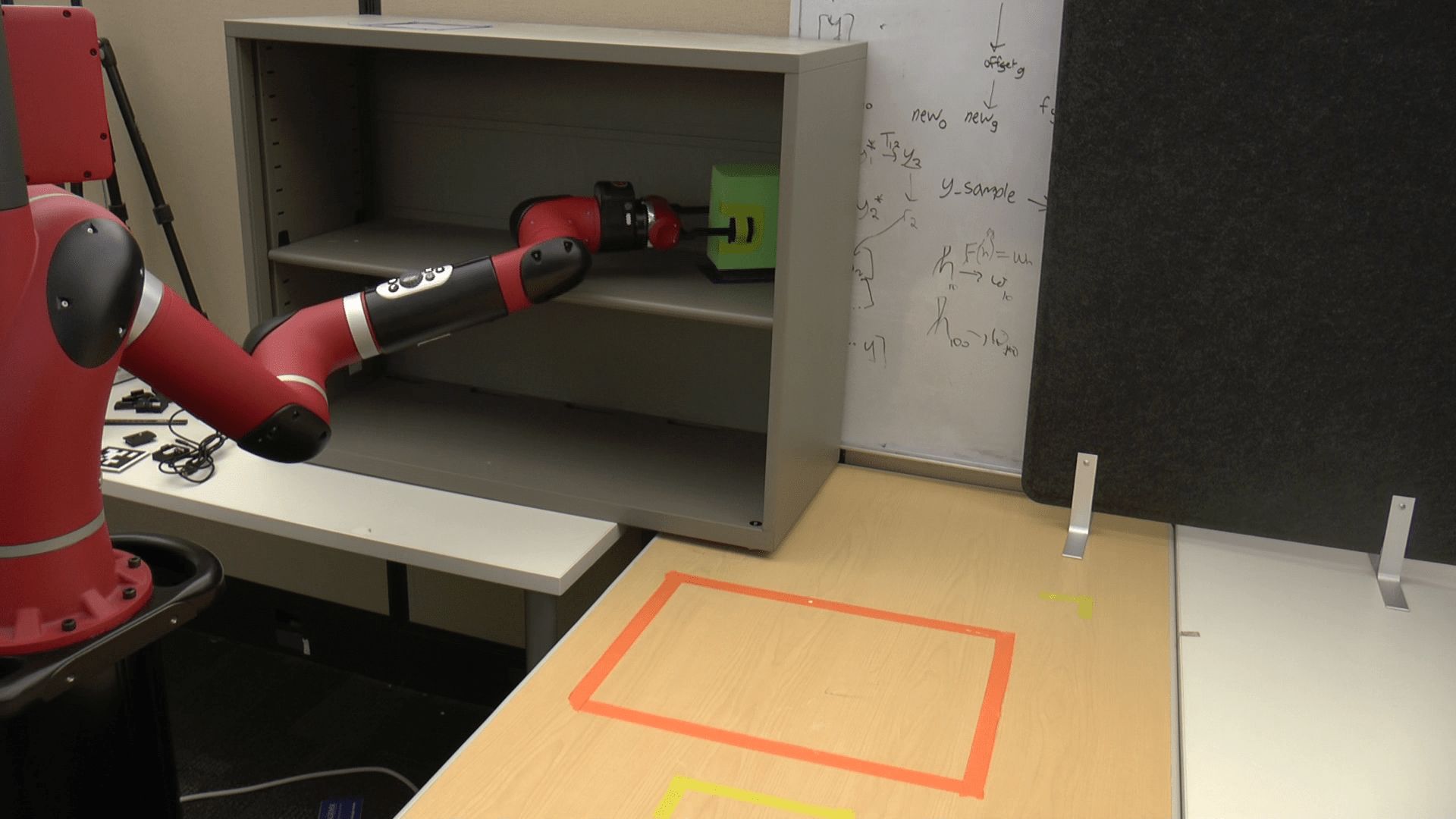}
   	\end{subfigure}
   	 \begin{subfigure}{1.\linewidth}	
	 	\includegraphics[trim=40 0 20 0, clip, width=\linewidth]{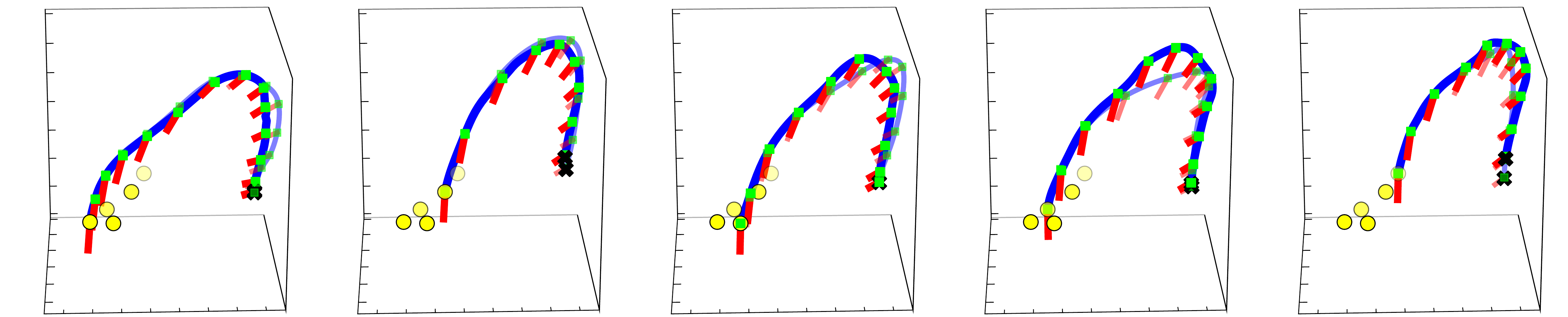}
   	\end{subfigure}
\caption{\small{The \emph{placing-2} task required the robot pick an object from a table, significantly rotate its end-effector, and place the object on a shelf. \emph{Top:} A series of snapshots showing a robot executing learned behavior. \emph{Bottom:} Plots of a subset of motion reproductions from different initial poses, overlaid on corresponding demonstrations. The yellow circles represent the initial end-effector positions, each corresponding to one of the rollouts. Note that the viewing angle in the plots is different from that in the robot execution snapshots.}}
\label{fig:placing_2}
\end{figure}

We evaluated our approach on three manipulations tasks\footnote{Accompanying video is available at: \texttt{\url{https://youtu.be/hwcxzLnxZPQ}}. } on a 7-DOF Rethink Sawyer robot with configuration space coordinates $\q\in\mathbb{R}^7$. We consider $3$ tasks including \emph{inspection}, \emph{placing-1}, and \emph{placing-2}. For details about the task specifications, the reader is referred to Figs.~\ref{fig:inspection}--\ref{fig:placing_2}. For each task, a human subject provided multiple configuration space demonstrations via kinesthetic teaching: $14$ demonstrations for \emph{inspection}, $9$ for \emph{placing-1}, and $12$ for \emph{placing-2}. 

 \begin{figure*}
 \centering
  	\begin{subfigure}{0.32\linewidth}
 	\centering
 	\includegraphics[trim=58 38 85 35, clip, width=1\linewidth]{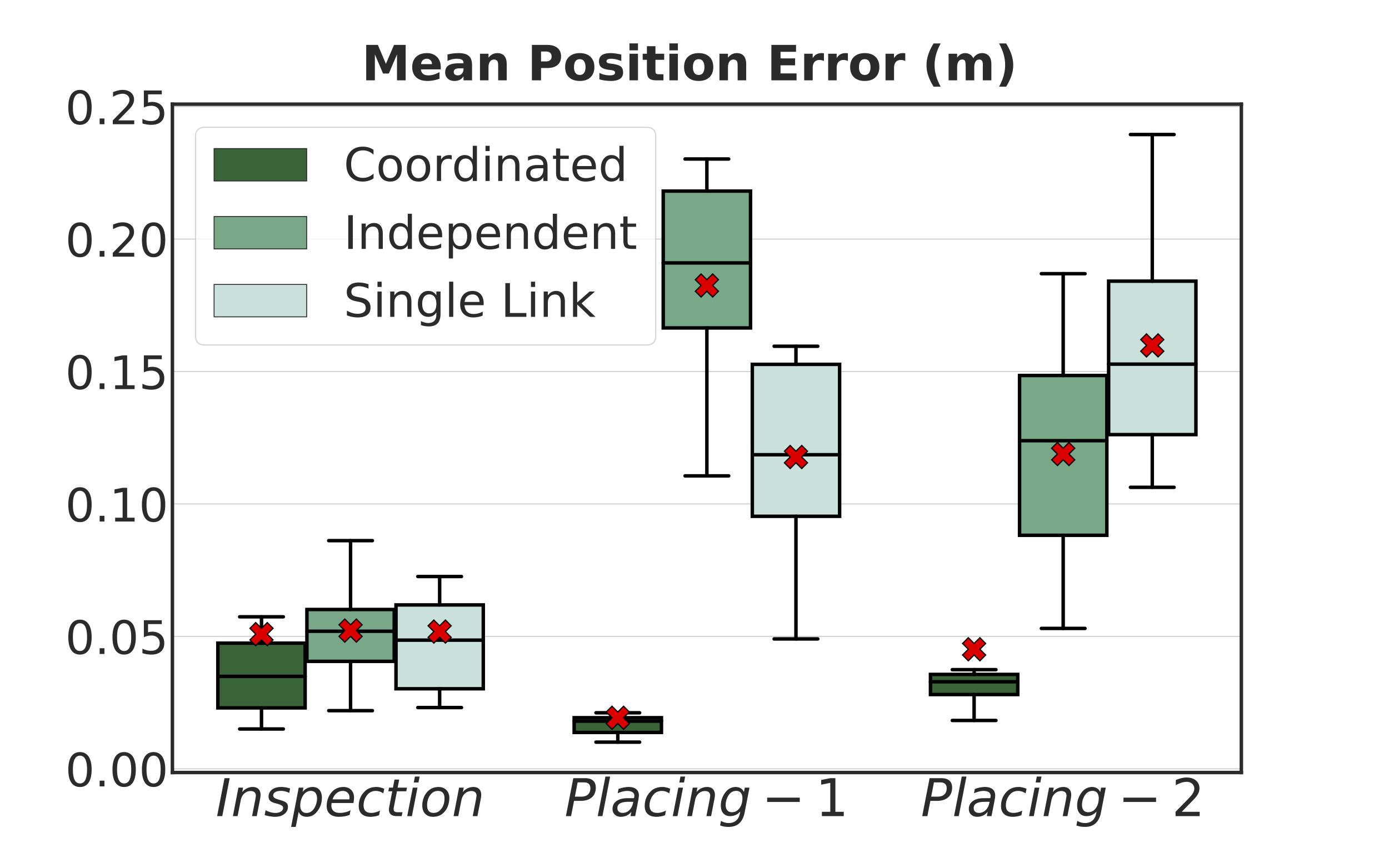}
 	\end{subfigure}
 	\begin{subfigure}{0.32\linewidth}
 	\includegraphics[trim=58 38 85 35, clip, width=1\linewidth]{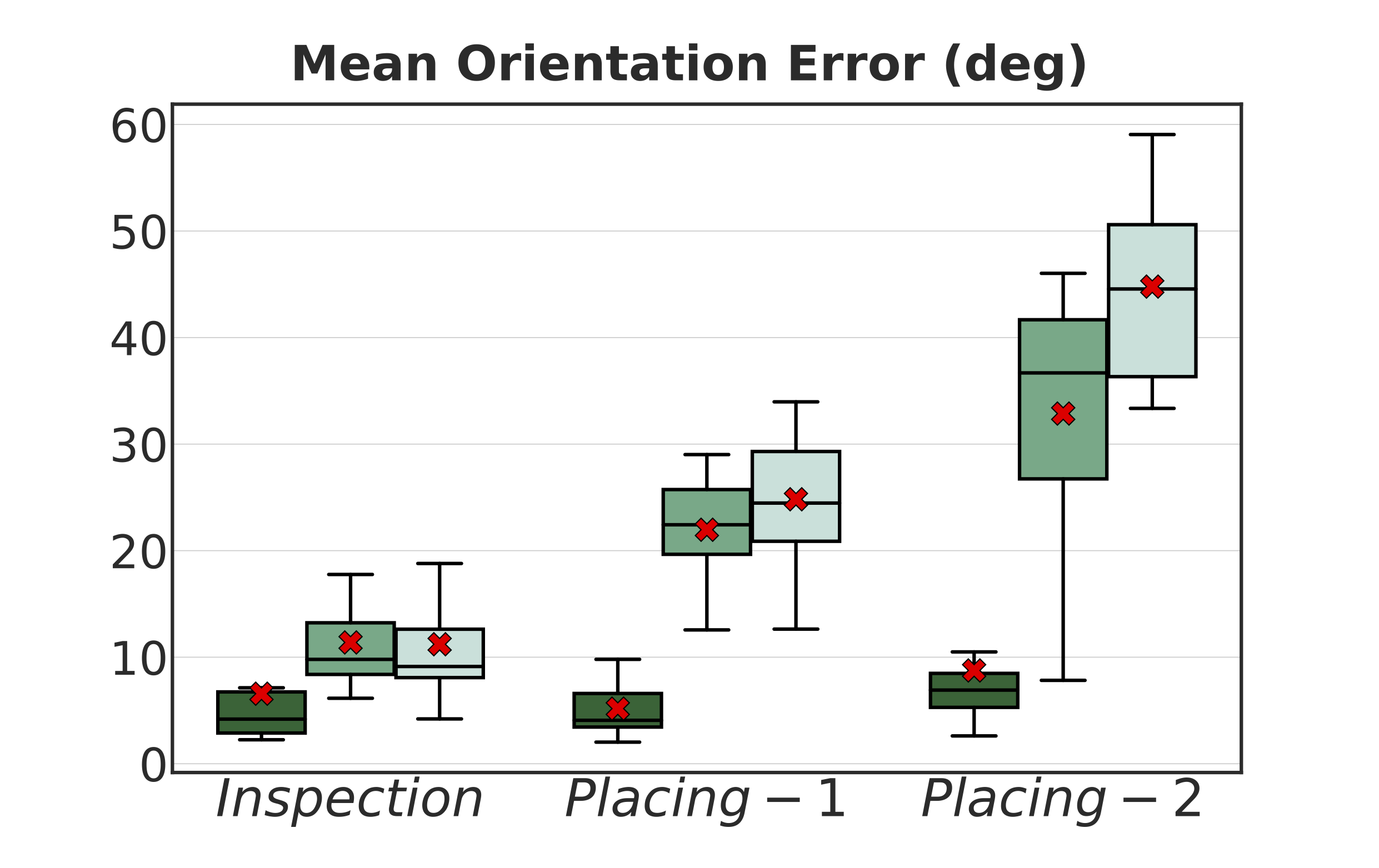} 
 	\end{subfigure}
  	\begin{subfigure}{0.32\linewidth}
 	\includegraphics[trim=58 38 85 35, clip, width=1\linewidth]{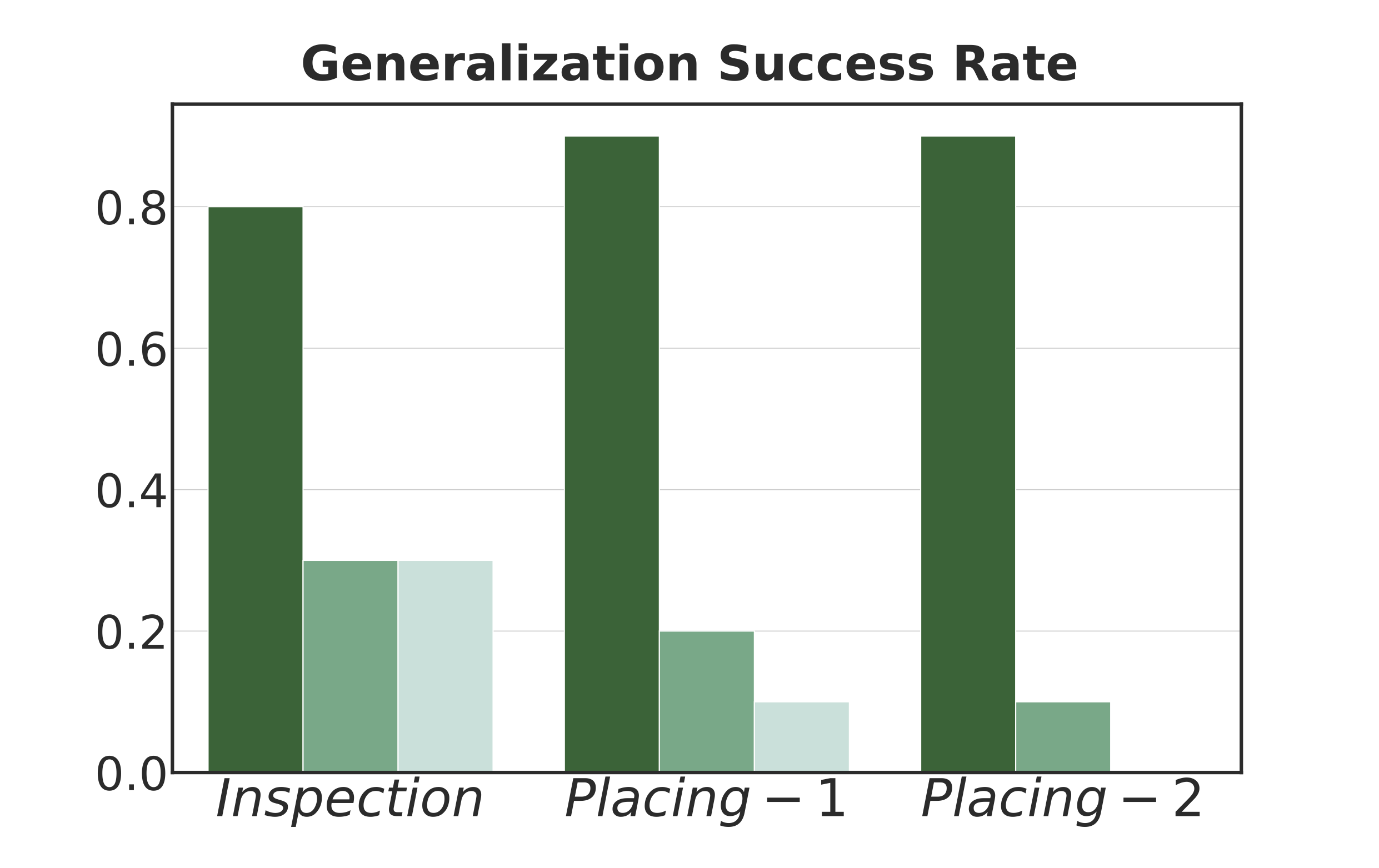}
 	\end{subfigure}
  	\caption{\small{Comparison of our approach against the baselines based on, mean position error (\emph{left}), mean orientation error (\emph{middle}), and generalization success rate over 10 executions (\emph{right}).}}
  \label{fig:comparison_quantitative}
 \end{figure*}
 
  \begin{figure*}
  \centering
  	\begin{subfigure}{0.15\linewidth}
  	\centering
  	\fbox{
 		\includegraphics[trim=0 0 0 0, clip, height=80px]{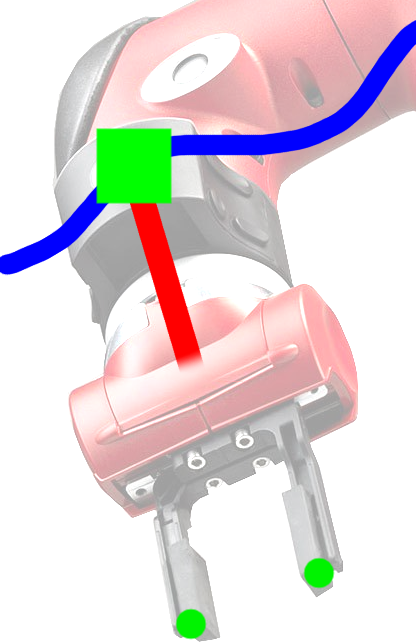}}
 	\end{subfigure}
 	\qquad
 	\fbox{\hspace{-0.5em}
 	\begin{subfigure}{0.7\linewidth}
 		\includegraphics[trim=80 10 80 0, clip, height=120px]{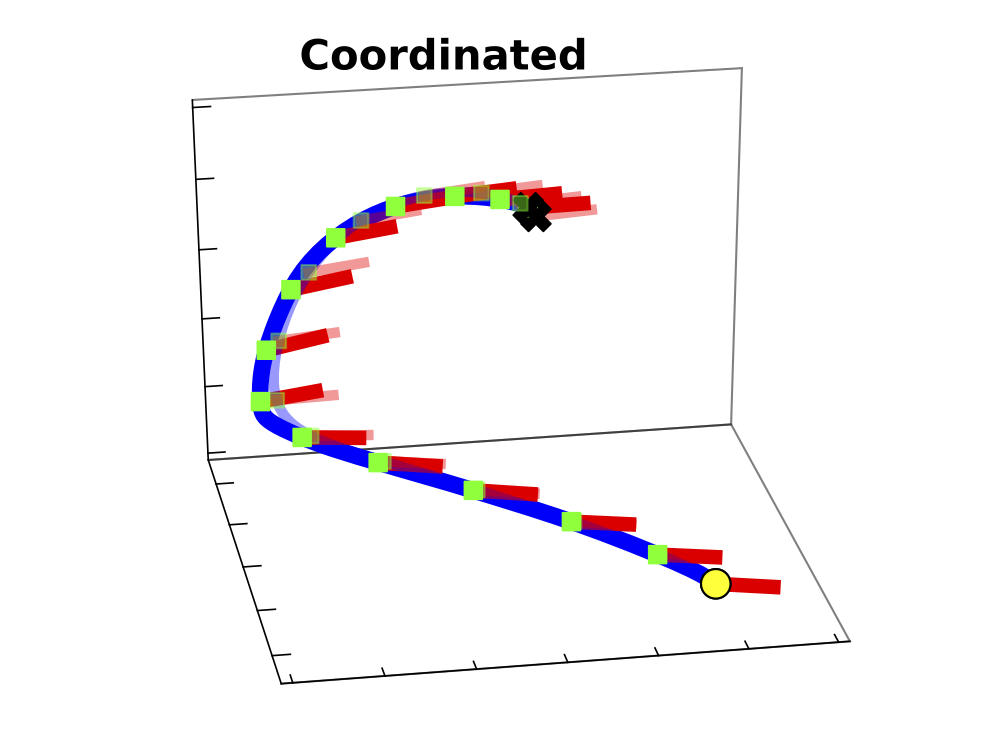}
  		\includegraphics[trim=80 10 80 0, clip, height=120px]{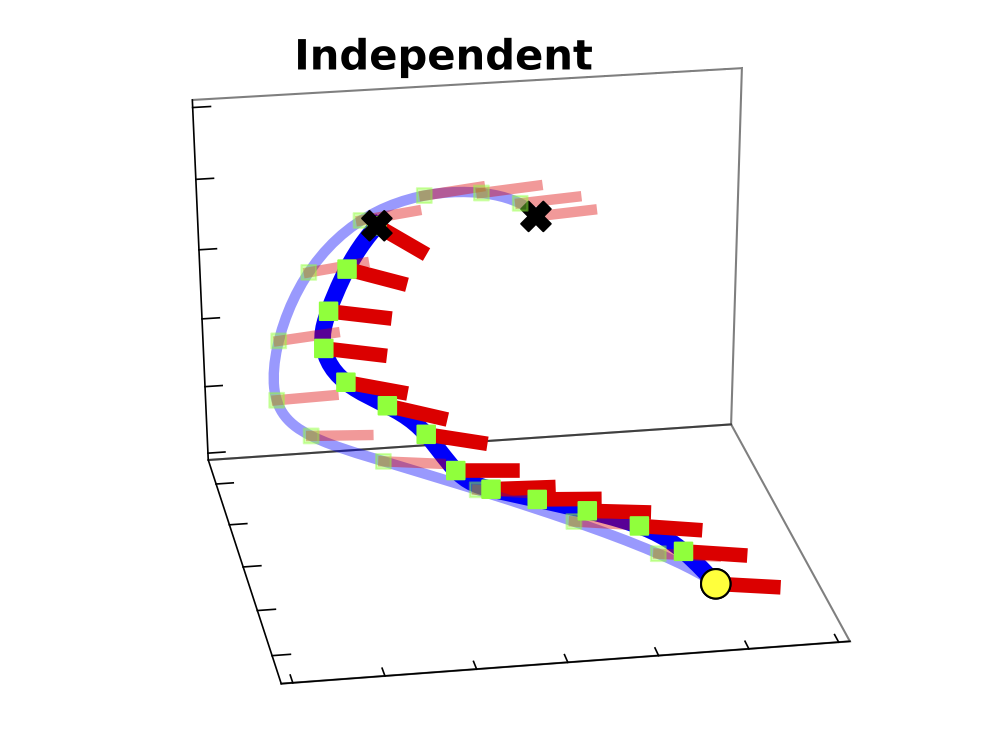}
  	 	\includegraphics[trim=80 10 80 0, clip, height=120px]{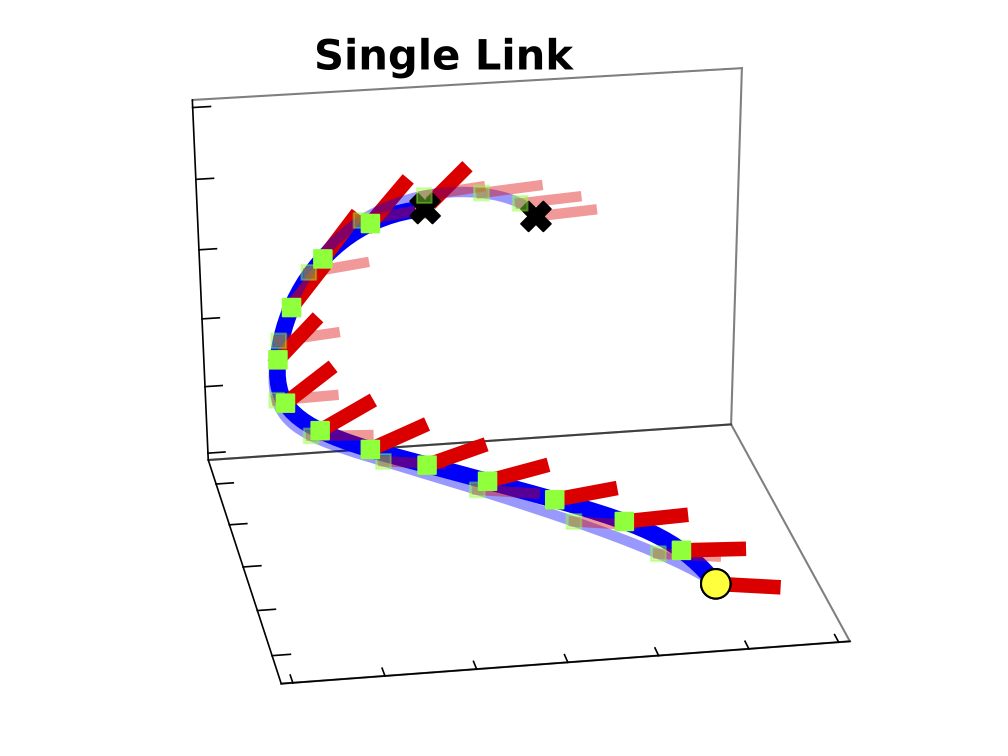}
 	\end{subfigure}\hspace{-0.5em}}
 	\caption{\small{\emph{Left}: Visualization of the $3$ control points (in green), with the end-effector control point denoted by a square while the two control points for gripper tips are given by circles. Overlaid is an end-effector position trajectory (in blue), and a line directed from the end-effector to the center of the gripper (in red) denoting instantaenous end-effector orientation. \emph{Right}: Plots showing example pose trajectories starting from an initial end-effector pose (yellow circle) governed by our approach and the baselines. Also shown in the background, is the demonstration starting from the same initial pose. The final positions are denoted by black crosses.}}
 	\label{fig:comparison_qualitative}
 \end{figure*}

\textbf{Subtasks: }
Each of the tasks is decomposed into learnable subtasks assigned to $3$ robot body parts, whereby each body part is represented by a unique \emph{control point} (see Fig.~\ref{fig:comparison_qualitative} for details). Given our choice of control points, the subtask policies effectively control the end-effector pose (i.e. position and orientation). However, we stress that our learning approach is not only limited to learning robot poses. In fact, one may instead, for instance, choose to learn motion policies dictating a partial pose (by removing a control point), or pose alongside the robot elbow (by adding an additional control point). 
Furthermore, there is hand-specified a default subtask policy pulling the end-effector in straight-line towards a desired goal pose. It is governed by a convex potential and a constant inertia matrix $\M = 10\mathbf{I}$. Additionally, to ensure the root importance weight matrix $\M_\rtt$ is always non-singular, we add a small offset $\epsilon_\rtt = 0.02$ to its diagonal entries. 

\textbf{Baselines: }
To evaluate the performance of our approach, we establish two baselines: \textit{(i)} an \emph{independent} learning version whereby the subtask policies are learned independently, which reproduces the setup of~\cite{rana2020learning}, and \textit{(ii)} a \emph{single link} learning version where just a single control point (i.e. end-effector) is chosen and the associated subtask policy is again learned independently.

\textbf{Learning Details: }
As is introduced in~Section~\ref{sec:approach}, the subtask policies are defined by a set of diffeomorphisms $\{\phi_k^{\theta_k}\}_{k=1}^{3}$ and a set of latent importance weight matrices $\{\M_{\dtt_k}^{\theta_k}\}_{k=1}^{3}$. In our parameterization, each diffeomorphism is composed of $M=10$ chained diffeomorphisms, each parameterized by $D=200$ random Fourier features with length-scale $l=0.45$. On the other hand, each latent importance weight matrix $\M_{\dtt_k}$ has two hidden layers with $128$ and $64$ dimensions respectively. The optimization problem in~\eqref{eq:learning_problem} was solved with Adam optimizer~\cite{kingma2014adam} with a learning rate of $1\times10^{-4}$ and weight decay $1\times10^{-8}$. 

\textbf{Results}
 Fig.~\ref{fig:comparison_qualitative} shows example reproductions of end-effector pose trajectories under the aforementioned variants of our algorithm. Our coordinated learning approach is observed to successfully reproduce the demonstrated motions. However, the baselines either fail to reproduce the position profile or the orientations. To quantitatively evaluate the capacity of our approach to reproduce demonstrations, we employ two error metrics i.e. mean position error, and mean orientation error. We evaluate position errors in terms of Euclidean distance i.e. $error(\bm{p}_1(t), \bm{p_2}(t)) = \|\bm{p}_1(t) - \bm{p}_2(t)\|_2$, where $\pb_1(t)$ and $\pb_2(t)$ are end-effector positions on the demonstrated and reproduced trajectory at time stamp $t$, respectively. On the other hand, for orientation error we employ $error(\bm{o}_1(t), \bm{o}_2(t)) = \arccos(|\bm{o}_1(t)\cdot\bm{o}_2(t)|)$, where $\bm{o_1}(t)$ and $\bm{o}_2(t)$ are unit quaternions representing end-effector orientations. For each comparison metric, we take the mean of the errors accumulated over the duration of a trajectory. Fig.~\ref{fig:comparison_quantitative} reports these comparisons as box plots. For the two \emph{placing} tasks, our approach outperforms the baselines by a significant margin. A major contributor towards this difference in performance is the existence of default subtask policies. When learned without accounting for the existing policies, the learned policies may not be able to sufficiently bias against the default behavior. Furthermore, we also observe that the independent learning version occasionally performs worse than the single link learning variant. This is perhaps because the independently learned subtask policies may conflict with each other too. This does not manifest as much in the single link case, since there is only one learned subtask policy. 

Lastly, we also test the generalization performance of our approach. For this evaluation, we execute rollouts of our motion policy from $10$ novel initial configurations. We considered a rollout as successful if all the goals of the task were met without any collisions. Fig.~\ref{fig:comparison_quantitative}(\emph{right}) reports the success rates. Once again, our end-to-end learning approach is seen to outperform the baselines in terms of generalization success rates. We also observed that while the difference in terms of quantitative errors between our approach and the baselines was small on the \emph{inspection} task, there were vast differences in performances given by generalization success rates. This is perhaps because, even when not trained end-to-end, the robot's kinematic constraints may enforce certain level of coordination between subtasks, thus resulting in low reproduction errors. However, for highly constrained tasks like the ones we explore in this paper, even small errors can result in task execution failures.
Rollouts from our learned motion policies, starting from the same configurations as demonstrations, are also visualized in Figs.~\ref{fig:inspection}--\ref{fig:placing_2} (\emph{bottom}). While we validated our approach on all the demonstrations, only a subset of rollouts is visualized here.

\addtolength{\textheight}{-4cm}   

\bibliographystyle{ieeetr}
\bibliography{refs}

\end{document}